%% file: main.tex
\PassOptionsToPackage{hyphens}{url}
\documentclass[sigconf, nonacm]{acmart}

\renewcommand\footnotetextcopyrightpermission[1]{}
\usepackage{booktabs}
\usepackage{amsmath}

\usepackage{amssymb}
\usepackage{graphicx}
\usepackage{listings}
\usepackage{xcolor}
\usepackage{multirow}
\usepackage{makecell}
\usepackage{enumitem}
\usepackage{algorithm}
\usepackage{algpseudocode}
\usepackage{textcomp}
\usepackage{microtype}
\usepackage{subcaption}
\usepackage{tabularx}
\usepackage{adjustbox}
\usepackage{tikz}
\usepackage{pgfplots}
\pgfplotsset{compat=1.18}

\definecolor{codegray}{rgb}{0.95,0.95,0.95}
\definecolor{codeblue}{rgb}{0.2,0.2,0.6}
\definecolor{codegreen}{rgb}{0.0,0.5,0.0}
\lstdefinestyle{nano}{
  backgroundcolor=\color{codegray},
  commentstyle=\color{codegreen}\itshape,
  keywordstyle=\color{codeblue}\bfseries,
  basicstyle=\ttfamily\footnotesize,
  breakatwhitespace=false,
  breaklines=true,
  captionpos=b,
  keepspaces=true,
  numbers=none,
  showspaces=false,
  showstringspaces=false,
  tabsize=2,
  frame=single,
  framerule=0.3pt
}
\acmConference[Preprint]{Preprint}{2026}{}
\acmYear{2026}
\copyrightyear{2026}
\acmISBN{}
\acmDOI{}

\begin{document}


\title[VectraYX-Nano]{VectraYX-Nano: A 42M-Parameter Spanish Cybersecurity Language Model with Curriculum Learning and Native Tool Use}

\author{Juan S. Santillana}
\affiliation{%
  \institution{Globant}
  \city{}
  \country{}}
\email{juan.salas@globant.com}


\begin{abstract}
We present \textsc{VectraYX-Nano}, a 41.95M-parameter decoder-only language model trained from scratch in Spanish for cybersecurity, with a Latin-American regional focus and native tool invocation via the Model Context Protocol (MCP). The model is built around four contributions. \textbf{(i)~Corpus.} \textsc{VectraYX-Sec-ES}, a 170M-token Spanish corpus assembled by an eight-VM distributed pipeline at $\sim$\$25 USD of cloud compute and partitioned into three curriculum phases: conversational (42M~tokens, OpenSubtitles-ES~\cite{lison2016opensubtitles} and OASST1~\cite{kopf2023openassistant}), cybersecurity (118M~tokens, NVD~\cite{nvd}, Wikipedia-ES, in-house NVD-derived Spanish CVE mirror, security blogs), and offensive-security tooling (10M~tokens, ExploitDB, HackTricks, OWASP). \textbf{(ii)~Architecture.} A 42M-parameter Transformer decoder combining Grouped-Query Attention~\cite{ainslie2023gqa}, QK-Norm~\cite{dehghani2023scaling}, RMSNorm~\cite{zhang2019rmsnorm}, SwiGLU~\cite{shazeer2020glu}, RoPE~\cite{su2024roformer}, and a $z$-loss auxiliary~\cite{chowdhery2023palm}, paired with a domain-balanced 16{,}384-token byte-fallback BPE~\cite{sennrich2016bpe, kudo2018sentencepiece} trained on a 50/50 conversational/technical mixture. \textbf{(iii)~Curriculum with replay.} Continual pre-training across the three phases with a replay buffer~\cite{ibrahim2024simple} mitigates catastrophic forgetting~\cite{french1999catastrophic, kirkpatrick2017overcoming} and yields a monotonic loss descent ($9.80 \!\to\! 3.17 \!\to\! 3.00 \!\to\! 2.16$). After SFT (final loss 1.74) on a curriculum-aware mixture of OASST-ES, Alpaca-ES, CVE Q\&A, and tool-use traces, the v2 bootstrap-ablation reference attains a conversational gate of $0.775 \pm 0.043$ on B5 over $N=4$ seeds (Section~\ref{sec:eval:multiseed}), and a controlled Phase-2 replay sweep over $\{0, 5, 10, 25, 50\}\%$ saturates B5 at $\geq 25\%$ replay (Section~\ref{sec:training:replay_sweep}). \textbf{(iv)~Two empirical findings, both N=4.} A controlled bootstrap-corpus ablation across v2 (OpenSubs), v4 (mC4-ES~\cite{xue2021mt5}), and v6 (60/25/15 OpenSubs/mC4/Wiki) under $N=4$ seeds exposes a \emph{loss-versus-register inversion}: the lower-perplexity bootstraps yield measurably worse conversational behavior ($v2 > v4 > v6$ on B5 at every paired seed), indicating that at the nano scale the bootstrap-corpus register dominates downstream chat quality. The B4 (tool-selection) floor of $0.000$ on the mixed SFT corpus is a \emph{corpus-density artifact}, not a capacity gate: rebalancing the SFT mixture to tool-use ratio 1:21 yields \textsc{VectraYX-Nano v7}, the released headline configuration, which reaches B4~$=0.230 \pm 0.052$ at 42M parameters ($N=4$ seeds) while retaining B1~$=0.332 \pm 0.005$ and B5~$=0.725 \pm 0.130$ (Section~\ref{sec:eval:headline_v7}); a LoRA~\cite{hu2022lora} replication on a 260M from-scratch mid-tier reaches $0.445 \pm 0.201$. The released GGUF~\cite{ggml2024gguf} artifact is 81~MB in F16 (approximately 20~MB in 4-bit quantization), runs at sub-second time-to-first-token on commodity hardware under \texttt{llama.cpp}~\cite{llamacpp}, and is, to the best of our knowledge, the first published Spanish-native cybersecurity LLM with end-to-end MCP integration. We release the corpus construction recipe, training scripts, configurations, GGUF weights, and the B1--B5 benchmark suite (B1:~500, B2:~200, B3:~100, B4:~200, B5:~314 prompts) for reproducibility.
\end{abstract}

\ccsdesc[500]{Computing methodologies~Natural language generation}
\ccsdesc[500]{Security and privacy~Software and application security}
\ccsdesc[300]{Computing methodologies~Neural networks}

\keywords{Language models, Cybersecurity, Spanish NLP, Curriculum learning, Tool use, Model Context Protocol, Edge inference}

\maketitle


\input{sections/01_introduction}
\input{sections/02_related_work}
\input{sections/03_corpus}
\input{sections/04_tokenizer}
\input{sections/05_architecture}
\input{sections/06_training}
\input{sections/07_tool_use}
\input{sections/08_evaluation}
\input{sections/08a_safety_evaluation}
\input{sections/09_discussion}
\input{sections/10_limitations}
\input{sections/11_conclusion}
\input{sections/12_next_steps}


\bibliographystyle{ACM-Reference-Format}
\bibliography{references}

\end{document}

%% file: sections/01_introduction.tex
\section{Introduction}
\label{sec:introduction}

Large language models (LLMs) have become a foundational tool for security analysts: they assist in vulnerability triage, log analysis, malware classification, and incident response. However, the publicly available LLM ecosystem suffers from two well-documented coverage gaps that compound when intersected. First, the strongest open-weight chat models are trained predominantly on English text~\cite{touvron2023llama, dubey2024llama3, qwen2024qwen25}, with Spanish typically representing a small fraction of pre-training mixtures, despite Spanish being the second-most-spoken native language in the world~\cite{ethnologue}. Second, while there is a growing literature on cybersecurity-specialized language models, virtually all of these models are trained on English corpora~\cite{aghaei2022securebert, bayer2024cysecbert} and none, to our knowledge, target Latin-American security terminology, regional CSIRT vocabularies (CCN-CERT, INCIBE, CSIRT-CL, COLCERT), or the LATAM threat-intelligence context.

These two gaps are jointly painful for security operations centers (SOCs) in Latin America. Spanish-speaking analysts who would otherwise benefit most from LLM assistance must work either with English-only domain models, with general-purpose Spanish models that lack technical accuracy, or with frontier closed-source models whose behavior they cannot audit, retrain, or deploy on-premise. The on-premise constraint is not academic: LATAM security teams routinely process classified incident reports, customer PII, and unreleased indicators of compromise that cannot leave the network.

A second motivation for this work is the rise of \emph{tool-augmented} language models~\cite{schick2023toolformer, qin2024toolllm, patil2024gorilla} and, more recently, the emergence of the Model Context Protocol (MCP)~\cite{anthropic2024mcp} as a standard for LLM--tool interfacing. Cybersecurity is one of the strongest application domains for tool use, since the underlying knowledge changes daily (new CVEs, KEV additions, TTP updates) and an analyst's typical query (``is this CVE being exploited?'', ``has this hash been flagged?'') has an authoritative external answer that a parametric model cannot reliably memorize. A small parametric model that knows \emph{when} to call a tool can be substantially more useful than a much larger model that hallucinates answers from a frozen training cutoff.

\paragraph{Contributions.} We present \textsc{VectraYX-Nano}, a 41.95M-parameter Spanish-language cybersecurity LLM trained from scratch with native MCP tool-use support. Our contributions are:

\begin{enumerate}[leftmargin=*]
\item \textbf{\textsc{VectraYX-Sec-ES} corpus.} We release the construction recipe for a 170M-token Spanish cybersecurity corpus assembled from a distributed pipeline of eight virtual machines. The corpus includes 88K NVD CVE entries, 50K previously translated Spanish CVEs from an in-house NVD-mirror SQLite store, a 53,590-article filtered Spanish Wikipedia subset (82M tokens, the single largest component), translated ExploitDB entries, the Spanish translations of HackTricks and OWASP, and a curated set of conversational Spanish from OpenSubtitles-ES~\cite{lison2016opensubtitles} and OASST1~\cite{kopf2023openassistant}. The full pipeline costs approximately \$25 USD in cloud compute.

\item \textbf{Modern small-LLM architecture.} We design a 41.95M-parameter Transformer decoder that integrates Grouped-Query Attention~\cite{ainslie2023gqa}, QK-Norm~\cite{dehghani2023scaling}, RMSNorm~\cite{zhang2019rmsnorm}, SwiGLU~\cite{shazeer2020glu}, RoPE~\cite{su2024roformer}, weight-tied embeddings, and a $z$-loss auxiliary~\cite{chowdhery2023palm}, alongside a domain-balanced 16{,}384-token byte-fallback BPE~\cite{sennrich2016bpe, kudo2018sentencepiece} tokenizer trained on a 50/50 conversational/technical mixture.

\item \textbf{Curriculum pre-training with replay.} We apply a three-phase curriculum (conversational~$\rightarrow$~cybersecurity~$\rightarrow$~tooling) with explicit replay buffers between phases following~\cite{ibrahim2024simple}. The phase weights are 100\% conversational $\rightarrow$ 75\%/25\% tech/conv $\rightarrow$ 70\%/20\%/10\% tools/tech/conv. The pre-training loss decreases monotonically across phases (9.80~$\rightarrow$~3.17~$\rightarrow$~3.00~$\rightarrow$~2.16) without observable catastrophic forgetting~\cite{french1999catastrophic, kirkpatrick2017overcoming}.

\item \textbf{Tool-use supervision via MCP.} We construct a 6{,}327-example tool-use SFT dataset templated against a real on-premise CVE database (50K Spanish CVEs, 27K exploits, 98K IOCs) and bound to six MCP servers (NVD, CISA KEV, MITRE ATT\&CK, OTX, LATAM intel, bash execution). The model learns to emit grammatically correct \texttt{<|tool\_call|>} JSON segments that the MCP runtime executes verbatim.

\item \textbf{Curriculum ablation: bootstrap-corpus register matters ($N=4$).} We report a controlled ablation that swaps OpenSubtitles-ES (Phase~1, v2) for mC4-ES~\cite{xue2021mt5} filtered with FineWeb-2~\cite{penedo2024fineweb} quality scores (Phase~1, v4), and for a 60/25/15 mixture of OpenSubs / mC4-ES / Wikipedia-ES (v6). All three configurations are retrained from scratch under four independent seeds ($\{42, 7, 13, 23\}$). The mC4-ES variant achieves consistently \emph{lower} loss in every subsequent phase ($-0.29$ in Phase~2, $-0.28$ in Phase~3, $-0.17$ in SFT) but consistently \emph{worse} conversational behavior on the B5 held-out chat gate: $0.775 \pm 0.043$ (v2) vs.\ $0.700 \pm 0.122$ (v4) vs.\ $0.675 \pm 0.083$ (v6). The $v2 > v4 > v6$ ordering is preserved at every paired seed comparison. We attribute this inversion to a register-mismatch effect: at the 42M-parameter scale, the bootstrap corpus dictates the model's default response style, and an encyclopedic web register cannot be reliably overwritten by SFT alone.

\item \textbf{Tool-use density threshold; \textsc{VectraYX-Nano v7} as the released headline model.} The B4 (tool-selection) floor of $0.000$ across all three bootstrap configurations of the mixed-SFT v2 reference is a corpus-density artifact, not a capacity gate. Rebalancing the SFT mixture to a tool-use ratio of $1{:}21$ yields \textsc{VectraYX-Nano v7}, the released headline configuration, which reaches B4~$=0.230 \pm 0.052$ at 42M parameters under $N=4$ seeds while retaining B1~$=0.332 \pm 0.005$ and B5~$=0.725 \pm 0.130$. A separate LoRA~\cite{hu2022lora} study reproduces the density threshold at $1{:}211 \to 1{:}21$: at $1{:}21$ a rank-16 LoRA reaches $0.145 \pm 0.046$ on Nano and $0.445 \pm 0.201$ on a 260M from-scratch mid-tier we call \textsc{VectraYX-Base} ($N{=}4$ seeds each). The capacity gate is therefore a first-token prior conflict, not a parametric limit.

\item \textbf{Post-Chinchilla compute ablation (v8--v15).} Extending pre-training from $\sim$850M tokens (v7, $\sim$0.49$\times$ Chinchilla-optimal) to $\sim$1.7B tokens ($\sim$2$\times$ Chinchilla-optimal) improves B1 modestly ($0.332 \to 0.341$) and preserves B5, but causes B4 to collapse to $0.000$ for all SFT recipes we tested except a long-SFT (6 epochs) curated-corpus configuration (v14) that recovers $B4 = 0.160$. None of the eight post-Chinchilla configurations strictly dominates v7 on (B1, B4, B5) jointly, so v7 remains the released model. We interpret this as evidence that tool-use \emph{at this parameter scale} is jointly gated by SFT corpus density \emph{and} pre-training compute budget, with the latter destroying tool-call plasticity once weight norms consolidate.

\item \textbf{Edge-deployable artifact.} We export the fine-tuned model to GGUF~\cite{ggml2024gguf} (F16: 81~MB; Q4: $\sim$20~MB), runnable under Ollama~\cite{ollama} or \texttt{llama.cpp}~\cite{llamacpp} with sub-second time-to-first-token on a Raspberry~Pi~4. The artifact includes weight-tied LM heads and the 25 reserved domain tokens.
\end{enumerate}

\paragraph{Scope.} VectraYX-Nano is positioned as a \emph{nano-scale} model: it is meant to assist analysts on edge devices and in air-gapped environments, not to compete with frontier 70B+ chat models on open-domain reasoning. Within its target envelope --- Spanish cybersecurity Q\&A, CVE summarization, threat classification, command completion, and tool dispatch --- we show that a careful corpus, a domain-balanced tokenizer, and curriculum pre-training with replay can extract qualitative behavior that a similarly sized monolithic pre-training run cannot.

\paragraph{Reproducibility.} All training scripts, configuration files, the curriculum sampler with replay-buffer code, the benchmark harness, the tool-use corpus, and the B1--B5 evaluation datasets are released at \url{https://github.com/vectrayx/vectrayx-nano-paper}. Model checkpoints and LoRA adapters are available at \url{https://huggingface.co/jsantillana/vectrayx-nano}. Section~\ref{sec:training} provides exact hyperparameters and Section~\ref{sec:evaluation} documents the held-out evaluation protocol. The corpus itself is partially released under upstream licenses (NVD, Wikipedia, ExploitDB, OpenSubtitles); the LATAM-curated portion is released as construction recipes rather than raw text, in line with current practice for security corpora.

%% file: sections/02_related_work.tex
\section{Related Work}
\label{sec:related-work}

\paragraph{Security-domain language models.} Domain-specialized models for security have a short but active history. SecureBERT~\cite{aghaei2022securebert} continues pre-training a RoBERTa~\cite{liu2019roberta} backbone on cybersecurity text and reports gains on entity recognition over generic BERT~\cite{devlin2019bert}. CySecBERT~\cite{bayer2024cysecbert} similarly continues from BERT on a 670K-document English security corpus and improves classification benchmarks. Earlier work in this line includes SciBERT~\cite{beltagy2019scibert}, which establishes the methodology of vocabulary-extending continual pre-training for technical domains. All of these models share two limitations from our perspective: they are encoder-only, and they are trained on English. We are not aware of any prior decoder-only generative model published with a Spanish cybersecurity specialization.

\paragraph{Spanish-language and multilingual models.} The Spanish NLP ecosystem has matured around BETO~\cite{canete2020beto} (a Spanish BERT), the RoBERTa-base-BNE family~\cite{gutierrez2022maria}, and more recently Salamandra~\cite{salamandra2024} from the Barcelona Supercomputing Center, an open Iberian decoder family. mC4~\cite{xue2021mt5} and CC-100~\cite{conneau2020unsupervised} have served as the standard Spanish pre-training corpora; FineWeb-2~\cite{penedo2024fineweb} is a more recent multilingual quality-filtered web release. These resources have unlocked general-purpose Spanish language modeling, but none of them targets a security domain or ships with a tool-use modality.

\paragraph{Small / nano-scale language models.} Several recent works show that careful data curation can produce competitive sub-billion-parameter models. SmolLM2-135M and SmolLM2-360M~\cite{allal2024smollm2} achieve strong nano-scale benchmarks via a recipe of high-quality web, code, and synthetic data. MobileLLM~\cite{liu2024mobilellm} establishes that depth is more useful than width at the sub-billion scale and that grouped attention combined with weight sharing is effective. TinyLlama~\cite{zhang2024tinyllama} reports that small models can be trained substantially past the Chinchilla~\cite{hoffmann2022chinchilla} optimum and continue to improve. We adopt several of these design intuitions (depth $\geq$ width, GQA, weight tying) and constrain ourselves to a single GPU training budget.

\paragraph{Tool-augmented and tool-using LLMs.} Toolformer~\cite{schick2023toolformer} introduced self-supervised insertion of tool calls; Gorilla~\cite{patil2024gorilla} demonstrated retrieval-anchored API calling at training time; ToolLLM~\cite{qin2024toolllm} curated 16K APIs with rich tool-use traces. More recently, Anthropic's Model Context Protocol (MCP)~\cite{anthropic2024mcp} has emerged as a standard for tool definitions and stateful tool sessions in production deployments. Tool-augmented work in security exists~\cite{deng2024pentestgpt}, but those systems orchestrate frontier models behind agentic loops; they do not train a small native tool-using model. To our knowledge, VectraYX-Nano is the first nano-scale Spanish security model with native tool-call generation evaluated against a held-out tool-selection benchmark.

\paragraph{Continual pre-training and replay.} Continual pre-training without replay tends to suffer from catastrophic forgetting of pre-training distribution, an issue documented since at least~\cite{french1999catastrophic} and given a Bayesian formulation in elastic weight consolidation~\cite{kirkpatrick2017overcoming}. Recent work on continual LLM pre-training~\cite{gupta2023continual, ibrahim2024simple} shows that simple strategies --- learning-rate re-warmup, modest replay percentages from the previous mixture, and adaptive token-budget allocation --- recover most of the lost performance with minimal overhead. We follow~\cite{ibrahim2024simple} closely and apply replay percentages of 25\% (Phase 2) and 10\% (Phase 3), which we find to control the loss trajectory but, as Section~\ref{sec:training} reports, materially affect the model's final stylistic register.

\paragraph{Curriculum learning.} Curriculum learning~\cite{bengio2009curriculum} predates the LLM era; it has been shown to help under specific data-quality regimes~\cite{soviany2022curriculum}. For LLM pre-training, ordering by difficulty or domain has had mixed results~\cite{xie2023doremi}, with the more reliable finding being that data \emph{mixture} matters more than data \emph{ordering}. Our setting is different: we are interested in instilling a stylistic register (chat-like Spanish) before specializing in a technical domain, and we report (Section~\ref{sec:training}) that for nano-scale models the ordering effect is large enough to flip user-visible behavior even when held-out perplexity moves the other way.

\paragraph{Edge-deployable language models.} On the deployment side, GGUF~\cite{ggml2024gguf} and \texttt{llama.cpp}~\cite{llamacpp} have made CPU inference of quantized small LLMs practical on commodity hardware including Raspberry Pi-class devices. Ollama~\cite{ollama} provides a developer-facing interface on top. We export to GGUF and ship a Modelfile so that VectraYX-Nano runs out of the box in this stack.

\paragraph{Positioning.} The intersection of (i) Spanish, (ii) cybersecurity, (iii) decoder-only / chat, (iv) trained from scratch, and (v) native MCP tool use --- to our knowledge --- is empty in the prior literature. The closest single-axis neighbors are SecureBERT~\cite{aghaei2022securebert} (security, English, encoder, no tools), Salamandra~\cite{salamandra2024} (Spanish, general-purpose, decoder, no tools), and Gorilla~\cite{patil2024gorilla} (English, general-purpose, decoder, tool use). VectraYX-Nano populates the intersection.

%% file: sections/03_corpus.tex
\section{The VectraYX-Sec-ES Corpus}
\label{sec:corpus}

\subsection{Design goals}
The corpus is designed to satisfy three constraints simultaneously: (i) sufficient Spanish-language conversational coverage to bootstrap chat behavior in a from-scratch model, (ii) substantive cybersecurity domain coverage in Spanish, including LATAM-specific vocabulary, and (iii) tool-use supervision grounded in a real CVE/IOC database rather than synthetic API descriptions. The corpus totals approximately 170M tokens after deduplication and is partitioned into three shards aligned with the curriculum phases: \texttt{phase1\_conv}, \texttt{phase2\_tech}, and \texttt{phase3\_tools}.

\subsection{Distributed collection pipeline}
The corpus was assembled by an eight-VM pipeline running in parallel for two days across two clouds (GCP and Azure), at an aggregate compute cost of approximately \$25 USD. The eight workers and their roles are:
\begin{itemize}[leftmargin=*]
\item \texttt{corpus-nvd}: NIST National Vulnerability Database REST API~\cite{nvd}; output 87,998 CVE records ($\sim$11.4M tokens).
\item \texttt{corpus-wiki}: MediaWiki API for the Spanish Wikipedia security categories.
\item \texttt{corpus-web}: 15 RSS feeds from Spanish-language security blogs.
\item \texttt{corpus-tech}: 7 RSS feeds from Spanish technology blogs.
\item \texttt{corpus-papers}: English security paper RSS feeds with Ollama-based translation~\cite{ollama}.
\item \texttt{corpus-malware}: Malpedia and MITRE ATT\&CK~\cite{strom2018mitre} surfaces with translation.
\item \texttt{corpus-exploitdb}: ExploitDB GitLab mirror with translation, yielding 2,385 translated entries.
\item \texttt{corpus-tools}: Eighteen GitHub repositories of pentesting tool documentation (Spanish branches).
\end{itemize}
On the central training node, additional sources are integrated locally: a 313~MB Wikipedia-ES dump processed via a streaming \texttt{bz2} + \texttt{iterparse} pipeline~\cite{wikipedia} (53,590 articles passed a 65-keyword cybersecurity filter, total $\sim$82M tokens, the single largest component of the corpus), GitHub clones of HackTricks-ES (952 documents, the \texttt{es} branch), HackTricks-Cloud (566), OWASP Top 10 (60), OWASP ASVS (42), OWASP WSTG (151), and PayloadsAllTheThings (139), and a previously assembled SQLite store hosted on an on-premise server we refer to internally as ``Pikachu''. Pikachu is not a third-party dataset: it is a long-running, locally maintained mirror of the NVD CVE database augmented with exploit, malware, and IOC tables, and an in-house Spanish translation pipeline that pre-translates each NVD entry as it is ingested. At the time of corpus assembly the Pikachu store contained 50,601 Spanish-translated CVEs (each derived from its NVD source record), 27,121 exploit entries, 7,556 malware signatures, and 98K IOCs.

\subsection{Translation policy}
For sources available only in English (papers, ExploitDB entries, malware reports), we translate locally using Ollama's \texttt{qwen2.5:1.5b}~\cite{qwen2024qwen25} with temperature $0.1$ and a 300-second timeout. We chose the 1.5B variant over the 3B variant after empirical comparison: 1.5B was approximately twice as fast and exhibited a lower timeout rate on long technical documents, with no observable degradation on chunked translations under 2{,}000 characters. Above this threshold we observed approximately 5--10\% mistranslation on highly technical text, and Section~\ref{sec:limitations} discusses the consequences.

\subsection{Phase composition}
\label{sec:corpus:phases}
Table~\ref{tab:corpus_phases} reports the per-phase composition. Tokens are reported under the final 16{,}384-vocabulary BPE tokenizer.

\begin{table}[t]
\caption{VectraYX-Sec-ES corpus by curriculum phase.}
\label{tab:corpus_phases}
\small
\begin{tabularx}{\columnwidth}{@{}llX@{}}
\toprule
\textbf{Phase} & \textbf{Tokens} & \textbf{Sources} \\
\midrule
\texttt{phase1\_conv}  & 42.4M  & OpenSubtitles-ES~\cite{lison2016opensubtitles}, OASST1-ES~\cite{kopf2023openassistant}, custom \\
\texttt{phase2\_tech}  & 117.7M & NVD~\cite{nvd}, Pikachu (NVD-mirror), Wikipedia-ES, blogs \\
\texttt{phase3\_tools} & 10.1M  & HackTricks-ES, OWASP-ES, ExploitDB, malware \\
\midrule
\textbf{Total}         & \textbf{170.2M} & \\
\bottomrule
\end{tabularx}
\end{table}

\subsection{Domain breakdown}
Table~\ref{tab:corpus_sources} provides a finer-grained breakdown by data source. A key observation is the dominance of the filtered Wikipedia-ES dump (82M tokens, $\sim$48\% of the corpus), which provides a broad foundation in Spanish technical writing across security, cryptography, malware, networking, and operating systems. The NVD CVE descriptions are bilingual (mostly English with some Spanish entries); the Pikachu store contributes 50K \emph{already translated} Spanish CVEs --- derived from the same NVD records but pre-translated by Pikachu's in-house ingestion pipeline --- that complement the raw NVD text.

\begin{table}[t]
\caption{Source-level breakdown of the pre-training corpus.}
\label{tab:corpus_sources}
\small
\setlength{\tabcolsep}{3pt}
\begin{tabularx}{\columnwidth}{@{}Xrrl@{}}
\toprule
\textbf{Source} & \textbf{Records} & \textbf{Tokens} & \textbf{Lang.} \\
\midrule
Wikipedia-ES (filtered)        & 53{,}590 & 82.0M & ES \\
NVD CVEs                       & 87{,}998 & 11.4M & ES/EN \\
Pikachu CVEs (NVD mirror, ES) & 50{,}601 &  8.8M & ES \\
Pikachu exploits               & 27,121 &  3.3M & EN/ES \\
GitHub repos (markdown ES)     &  1,884 &  3.7M & ES \\
Wikipedia-ES (API, security)   &     947  &  2.2M & ES \\
ExploitDB (translated)         &  2,385 &  1.3M & ES \\
Malware DBs (Malpedia, MITRE)  &     300  &  0.7M & ES \\
Pikachu malware signatures     &  7,556 &  0.6M & EN \\
Papers EN$\rightarrow$ES       &     236  &  0.6M & ES \\
Tech blogs ES                  &     322  &  0.5M & ES \\
Security blogs ES              &     241  &  0.4M & ES \\
Tools docs ES                  &     65+  &  0.3M & ES \\
\midrule
OpenSubtitles-ES               & 16,317 chunks & 39.0M & ES \\
OASST1-ES                      & 13,000 &  3.5M & ES \\
\bottomrule
\end{tabularx}
\end{table}

\subsection{Conversational subcorpus}
\label{sec:corpus:conv}
The conversational subcorpus is intentionally small (42M tokens, $\sim$25\% of the total). We assemble it from three sources: OpenSubtitles-ES from OPUS~\cite{lison2016opensubtitles} (16,317 chunks, $\sim$39M tokens of subtitle dialogue), the Spanish-filtered subset of OASST1~\cite{kopf2023openassistant} (13,000 conversations, $\sim$3.5M tokens), and 214 short hand-curated dialogues that establish cybersecurity-specific greetings and disclaimers. Section~\ref{sec:training} reports an ablation that swaps the OpenSubtitles component for filtered mC4-ES~\cite{xue2021mt5} and finds that, despite mC4-ES yielding lower overall pre-training loss, the OpenSubtitles bootstrap produces measurably better chat behavior.

\subsection{SFT corpus}
The SFT corpus is separate from pre-training. It contains 13K OASST1-ES conversations, 4{,}030 CVE Q\&A pairs derived from the Pikachu store, 6{,}327 tool-use traces (Section~\ref{sec:tooluse}), and 214 custom cybersecurity greetings, for approximately 28M tokens. In the family-scale extension (Section~\ref{sec:evaluation}), the Qwen2.5 fine-tuning corpus also includes \texttt{bertin-project/alpaca-spanish}~\cite{bertin2023alpaca} (51,942 examples) and the Spanish-filtered subset of OpenAssistant~OASST2~\cite{kopf2023openassistant} (18,022 conversations).

\subsection{Tokenization and binarization}
After training the BPE tokenizer (Section~\ref{sec:tokenizer}), we tokenize each phase into \texttt{uint16} memory-mapped binary shards in the style of nanoGPT~\cite{karpathy2023nanogpt}, with each phase occupying its own shard directory so that the curriculum sampler can read them at arbitrary mixture weights without re-tokenizing. The total compressed shard footprint is $\sim$340~MB. Phase~1 fits in a single shard; Phase~2 spans three shards; Phase~3 occupies one shard.

\subsection{Deduplication and quality filtering}
We deduplicate at the document level using exact-match hashing on a normalized form (lowercased, whitespace-collapsed). Cross-source deduplication is non-trivial when the same CVE description appears verbatim in NVD, in the Pikachu store, and in the Wikipedia-ES filtered dump; we tolerate this redundancy because the three surfaces correspond to different stylistic registers (terse advisory, translated narrative, encyclopedic prose) and we observe empirically that repeated exposure to the same vulnerability across these registers helps the model produce a more consistent response style at SFT time.

\subsection{Licensing and release}
The released portions of the corpus respect upstream licenses (CC0/CC-BY for Wikipedia, CC-BY-SA for HackTricks, public domain for NVD, CC-BY-NC for OpenSubtitles, Apache-2.0 for OASST1). The LATAM-curated portion that combines internal SQLite-derived translations with hand-curated dialogues is released as a construction recipe rather than raw text, in line with current security-corpus practice.

%% file: sections/04_tokenizer.tex
\section{Domain-Balanced Tokenizer}
\label{sec:tokenizer}

\subsection{Design rationale}
A first iteration of the tokenizer was trained on a 95\%-technical mixture and produced a vocabulary that fragmented common Spanish chat tokens (\textit{``hola''}, \textit{``gracias''}, \textit{``estás''}) while merging long technical n-grams. We replaced it with a 50/50 mixture for two reasons. First, the post-tokenization \emph{token budget} for chat sentences directly affects how many gradient updates a chat example contributes to the loss; oversplit chat tokens are a hidden source of register imbalance in the loss. Second, byte-fallback is necessary to guarantee that the model can serialize CVE identifiers, hashes, and base64 payloads without producing \texttt{<unk>} tokens.

\subsection{Configuration}
We use SentencePiece~\cite{kudo2018sentencepiece} BPE~\cite{sennrich2016bpe} with the configuration listed below, taken verbatim from \texttt{configs/nano.json}:
\begin{lstlisting}
{
  "vocab_size": 16384,
  "model_type": "bpe",
  "character_coverage": 1.0,
  "byte_fallback": true,
  "normalization": "nmt_nfkc",
  "split_digits": true,
  "split_by_unicode_script": true,
  "add_dummy_prefix": true,
  "balance": {
    "conversational_ratio": 0.5,
    "technical_ratio":      0.5
  }
}
\end{lstlisting}
The training corpus for the tokenizer is a 2M-line balanced sample drawn evenly from \texttt{phase1\_conv} (chunks from OpenSubtitles-ES and curated dialogues) and \texttt{phase2\_tech} + \texttt{phase3\_tools} (chunks from NVD, Wikipedia-ES, HackTricks-ES, ExploitDB, etc.). We use \texttt{nmt\_nfkc} normalization to preserve Spanish accents (NFKC without NFD decomposition); \texttt{split\_digits=true} ensures CVE numerics tokenize predictably; \texttt{byte\_fallback=true} guarantees full UTF-8 coverage.

\subsection{Special tokens}
The vocabulary reserves 27 user-defined symbols at the start, organized in five groups:
\begin{itemize}[leftmargin=*]
\item \textbf{Control}: \texttt{<|pad|> <|bos|> <|eos|> <|unk|> <|sep|>} (assigned to slots 0--4).
\item \textbf{Chat roles}: \texttt{<|system|> <|user|> <|assistant|> <|end|>}.
\item \textbf{Tool use}: \texttt{<|tool\_call|> <|/tool\_call|> <|tool\_result|> <|/tool\_result|>}.
\item \textbf{Domain}: \texttt{<|cve|> <|cvss|> <|ioc|> <|ttp|> <|mitre|> <|kev|> <|exploit|> <|patch|> <|alert|>}.
\item \textbf{Severity}: \texttt{<|critical|> <|high|> <|medium|> <|low|> <|info|>}.
\end{itemize}
Reserving these as user-defined symbols (rather than relying on the BPE merges to discover them) ensures that they always tokenize as exactly one token. The remaining 16{,}357 vocabulary slots are filled by BPE merges.

\subsection{Empirical token economy}
On Spanish chat sentences the v2 tokenizer is approximately twice as efficient as the prior v1 technical-only tokenizer. Representative examples (single-token decoding shown):
\begin{itemize}[leftmargin=*]
\item \texttt{"vulnerabilidad"} $\rightarrow$ 1 token (v1: 4--5 tokens).
\item \texttt{"CVE-2021-44228"} $\rightarrow$ 5 tokens (\texttt{CVE}, \texttt{-}, \texttt{2021}, \texttt{-}, \texttt{44228}); digit splitting keeps year and ID symbolic.
\item \texttt{"\textexclamdown Hola! \textquestiondown cómo estás?"} $\rightarrow$ 9 tokens (v1: $\sim$18 tokens).
\item \texttt{"<|user|>\textquestiondown qué es ransomware?<|end|>"} $\rightarrow$ 8 tokens (chat role markers as single tokens).
\end{itemize}
The asymmetric improvement on chat tokens (v1 nearly doubled) is the empirical effect that the design targets.

\subsection{Vocabulary size choice}
We chose 16{,}384 (rather than the more common 32K or 50K) on three grounds. (i) At 42M parameters with weight-tied embeddings, the embedding matrix is $16384 \times 512 = 8.4$M parameters, $\sim$20\% of the total budget. Doubling the vocabulary to 32K would push the embedding share to $\sim$33\%, leaving substantially less budget for transformer blocks. (ii) On the deployment side, 16K vocabularies quantize cleanly under GGUF Q4 schemes. (iii) Empirically, the 16K vocabulary covers the corpus with $<$0.05\% byte-fallback rate (i.e., almost all characters tokenize through merges rather than falling back to byte units), which we treat as a sufficient compression target for a domain model.

%% file: sections/05_architecture.tex
\section{Architecture}
\label{sec:architecture}

\subsection{Overall design}
VectraYX-Nano is a Transformer~\cite{vaswani2017attention} decoder-only language model with eight pre-norm blocks. The architecture follows the modern Llama-family stack~\cite{touvron2023llama, dubey2024llama3}: RMSNorm~\cite{zhang2019rmsnorm} for normalization, SwiGLU~\cite{shazeer2020glu} for the FFN nonlinearity, RoPE~\cite{su2024roformer} for positional encoding, weight-tied input/output embeddings, no biases on linear layers, and a $z$-loss auxiliary~\cite{chowdhery2023palm} on the logits. We add two stability improvements that have become standard in recent small-LLM releases: Grouped-Query Attention (GQA)~\cite{ainslie2023gqa} with $\textit{n}_q = 8$ and $\textit{n}_{kv} = 2$, and QK-Norm~\cite{dehghani2023scaling, henry2020querykey} that applies RMSNorm independently to query and key projections.

\subsection{Hyperparameters}
Table~\ref{tab:arch} reports the architecture configuration as drawn from \texttt{configs/nano.json}.

\begin{table}[t]
\caption{VectraYX-Nano architecture (\texttt{configs/nano.json}).}
\label{tab:arch}
\small
\begin{tabular}{lr}
\toprule
\textbf{Hyperparameter} & \textbf{Value} \\
\midrule
Vocabulary size           & 16{,}384 \\
Layers ($L$)              & 8 \\
Hidden dimension ($d_\mathrm{model}$) & 512 \\
Query heads ($n_q$)       & 8 \\
KV heads ($n_{kv}$)       & 2 \\
Head dimension ($d_h$)    & 64 \\
FFN dimension ($d_\mathrm{ffn}$) & 2{,}048 \\
Maximum sequence length   & 1{,}024 \\
RoPE base ($\theta$)      & 10{,}000 \\
RMSNorm $\epsilon$        & $10^{-6}$ \\
Init std ($\sigma$)       & 0.02 \\
Residual init scale       & $\sigma / \sqrt{2L} = 0.005$ \\
Dropout                   & 0.0 \\
Tied embeddings           & yes \\
QK-Norm                   & yes \\
$z$-loss coefficient      & $10^{-4}$ \\
\midrule
\textbf{Total parameters} & \textbf{41.95M} \\
\bottomrule
\end{tabular}
\end{table}

\subsection{Grouped-Query Attention}
With $n_q = 8$ and $n_{kv} = 2$, each KV head is shared by four query heads. The attention layer's projection footprint is therefore:
\begin{itemize}[leftmargin=*]
\item $W_Q \in \mathbb{R}^{512 \times 512}$ (8 query heads of dim 64),
\item $W_K, W_V \in \mathbb{R}^{512 \times 128}$ (2 KV heads of dim 64),
\item $W_O \in \mathbb{R}^{512 \times 512}$.
\end{itemize}
Compared to standard MHA at the same width ($W_K, W_V \in \mathbb{R}^{512 \times 512}$), GQA saves $\sim$50\% of the KV parameters and shrinks the KV cache by $4\times$ at inference time. We compute attention through PyTorch's \texttt{F.scaled\_dot\_product\_attention} with \texttt{is\_causal=True}, which dispatches to FlashAttention-2~\cite{dao2023flashattention2} on supported hardware (NVIDIA L4 in our case). Implementation details are visible in \texttt{model/transformer.py}: queries are projected to $(B, n_q, T, d_h)$, keys and values to $(B, n_{kv}, T, d_h)$, and KV is repeated $n_q / n_{kv} = 4$ times along the head dimension before the kernel call.

\subsection{QK-Norm}
We apply RMSNorm to the query and key projections after RoPE rotation but before the attention dot product. This stabilizes training at small batch sizes and small head counts: empirically the gradient norm trace is smoother and we observed no loss spikes during the 3{,}519 pre-training steps despite running BF16 without a loss scaler. QK-Norm has been adopted by OLMo~\cite{groeneveld2024olmo} and Gemma~\cite{team2024gemma} for similar reasons.

\subsection{Initialization}
We initialize linear and embedding weights from $\mathcal{N}(0, \sigma^2)$ with $\sigma=0.02$, and rescale residual-output projections (the \texttt{wo} of attention and the \texttt{w\_down} of SwiGLU) by $\sigma / \sqrt{2L}$ following the GPT-2 scaled init~\cite{radford2019gpt2}. This keeps the variance of activations bounded as the depth grows and is critical for stable BF16 training with $z$-loss.

\subsection{$z$-loss}
The PaLM $z$-loss auxiliary~\cite{chowdhery2023palm} regularizes the partition function of the softmax, $z = \log \sum_i \exp(\ell_i)$, by adding $\lambda \cdot \mathbb{E}[z^2]$ to the cross-entropy loss with $\lambda = 10^{-4}$. This term keeps the unnormalized logit magnitudes from drifting upward over long training runs and is a cheap insurance against the well-known instability of bf16 softmax over large vocabularies.

\subsection{Total parameter accounting}
The 41.95M parameter total decomposes approximately as: embedding/LM-head 8.4M (tied; counted once), 8 transformer blocks of $\sim$4.2M each ($\sim$33.5M) where each block contains $\sim$0.6M attention parameters and $\sim$3.1M FFN parameters, plus $\sim$30K parameters across the nine RMSNorm gain vectors. We confirmed the total at training startup; the script logs \texttt{[model] 41.95M params} at every run.

\subsection{Inference profile}
At inference, the GQA design and small context (1{,}024 tokens) keep the KV cache footprint to $L \cdot 2 \cdot n_{kv} \cdot d_h \cdot T \cdot 2\,\text{bytes (BF16)} = 8 \cdot 2 \cdot 2 \cdot 64 \cdot 1024 \cdot 2 = 4$~MiB per sequence. After GGUF Q4 quantization, weights occupy $\sim$20~MB. The combination yields a total resident set of $\sim$60--80~MB at inference, which fits comfortably in the L1+L2+L3 cache hierarchy of modern x86 CPUs and in the 1~GB RAM budget of a Raspberry~Pi~4. We measure 6--10 tokens/s on a Raspberry~Pi~4 (Cortex-A72, 4 cores) and 60--100 tokens/s on a contemporary laptop CPU.

\subsection{Why this configuration?}
Two configuration choices warrant explicit justification. (i) \emph{Depth $\geq$ width}. Following MobileLLM~\cite{liu2024mobilellm}, we prefer 8 layers of width 512 to 4 layers of width 1{,}024 at the same parameter budget, because depth empirically helps reasoning behavior more than width at sub-100M scales. (ii) \emph{Aggressive GQA ratio}. We use $n_q/n_{kv} = 4$, which is more aggressive than Llama-2-7B's $1\!:\!1$ but matches Mistral-7B and Llama-3.2 conventions. At our parameter scale the savings from GQA are absolute (4M parameters reclaimed for FFN width), and we did not observe any quality degradation on chat or CVE-Q\&A relative to a small ablation pilot run with $n_q/n_{kv} = 2$.

%% file: sections/06_training.tex
\section{Curriculum Pre-training with Replay}
\label{sec:training}

\subsection{Motivation: failures of monolithic pre-training (v1)}
Our first pre-training attempt (v1) followed the conventional small-LLM recipe: a single-phase pre-training over the full technical corpus (142M tokens, two epochs) followed by a multi-stage SFT. The resulting model reached pre-training loss 3.35 and SFT loss 0.315, yet exhibited a striking failure mode: it answered the prompt \texttt{"hola"} (Spanish for ``hello'') with a CVE analysis instead of a greeting. Three increasingly conversational SFT runs (v1: 86K examples, v2: 11K examples, v3: 24K examples mixing OASST1) failed to repair this behavior. We conclude that a model that has not seen sufficient conversational Spanish during pre-training cannot be coerced into chat behavior by SFT alone at the 42M scale --- the chat register has to be the model's first language.

\subsection{Three-phase curriculum}
\label{sec:training:curriculum}
We address this with a three-phase curriculum:
\begin{description}[leftmargin=0pt, font=\normalfont\itshape]
\item[Phase 1 -- conversational bootstrap.] 100\% \texttt{phase1\_conv} (42.4M tokens, 2 epochs). The model establishes a default Spanish chat register before encountering any technical text.
\item[Phase 2 -- domain immersion.] 75\% \texttt{phase2\_tech} + 25\% \texttt{phase1\_conv} replay (117.7M tokens). Continued pre-training resumed from the Phase 1 checkpoint at a lower learning rate. The 25\% replay buffer follows~\cite{ibrahim2024simple} and is intended to prevent forgetting of conversational Spanish while the model absorbs CVE/Wikipedia/blog text.
\item[Phase 3 -- tooling specialization.] 70\% \texttt{phase3\_tools} + 20\% \texttt{phase2\_tech} + 10\% \texttt{phase1\_conv} (10.1M tokens). A short, low-LR phase that pulls the model toward HackTricks, OWASP, and ExploitDB content with a smaller replay tail of both prior phases.
\end{description}
Implementation: a single \texttt{MixedCurriculumDataset} (Listing~\ref{lst:curriculum}) memory-maps three shard directories and samples each batch index from a categorical distribution determined by the phase weights. This formulation makes the replay percentage a single hyperparameter per phase rather than a separate dataset construction step.

\begin{lstlisting}[language=Python, caption={Replay-aware curriculum sampler (excerpt from \texttt{curriculum\_dataset.py}).}, label=lst:curriculum]
def make_phase_mix(phase_idx, replay_conv=None, replay_tech=None):
    if phase_idx == 1:
        return {"phase1_conv": 1.0, "phase2_tech": 0.0, "phase3_tools": 0.0}
    if phase_idx == 2:
        rc = 0.25 if replay_conv is None else replay_conv
        return {"phase1_conv": rc, "phase2_tech": 1.0 - rc, "phase3_tools": 0.0}
    if phase_idx == 3:
        rc = 0.10 if replay_conv is None else replay_conv
        rt = 0.20 if replay_tech is None else replay_tech
        return {"phase1_conv": rc, "phase2_tech": rt,
                "phase3_tools": 1.0 - rc - rt}
\end{lstlisting}

\begin{figure}[t]
\centering
\includegraphics[width=\columnwidth]{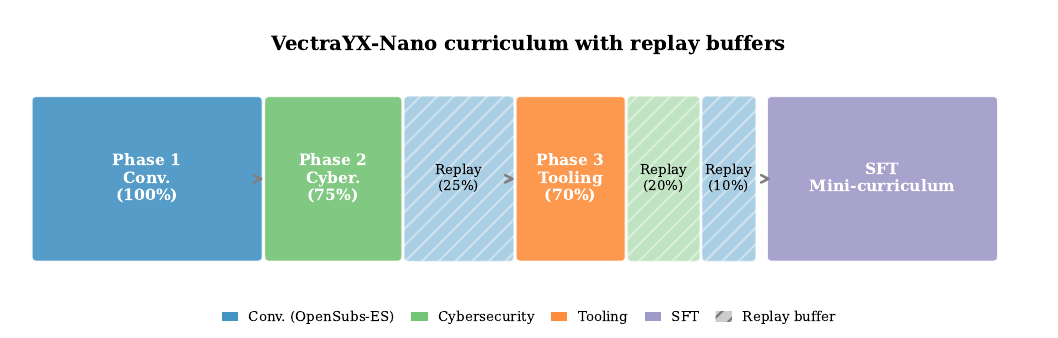}
\Description{Schematic of the three-phase pre-training curriculum. Phase 1 (conversational) is 100\% conversational data; Phase 2 (cybersecurity) is 75\% technical with 25\% conversational replay; Phase 3 (tooling) is 70\% tools, 20\% technical, and 10\% conversational replay.}
\caption{Three-phase curriculum with replay. Each phase samples from a weighted mixture of shard directories; the replay fractions (25\% in Phase~2, 10\%+20\% in Phase~3) prevent catastrophic forgetting of earlier registers.}
\label{fig:curriculum_schema}
\end{figure}

\subsection{Hyperparameters and infrastructure}
\label{sec:training:hyperparams}
Table~\ref{tab:training_hp} reports the per-phase optimizer schedule. All phases use AdamW~\cite{loshchilov2019adamw} with $\beta_1=0.9$, $\beta_2=0.95$, weight decay $0.1$ for pre-training, gradient clipping at $\|g\|_2 \leq 1.0$, BF16 mixed precision on an NVIDIA L4 GPU (23~GB VRAM, GCP \texttt{us-west1-a}), and a cosine learning-rate schedule with linear warmup. The effective tokens per optimizer step are $B \cdot G \cdot T = 16 \cdot 8 \cdot 1024 = 131{,}072$ during pre-training and $16 \cdot 4 \cdot 1024 = 65{,}536$ during SFT. We measured throughput between 40{,}847 and 40{,}883 tokens/second across all three pre-training phases (peak GPU utilization 99--100\%).

\begin{table}[t]
\caption{Training hyperparameters per phase. LR is the cosine peak; warmup is the fraction of total steps.}
\label{tab:training_hp}
\small
\setlength{\tabcolsep}{4pt}
\begin{tabular}{lrrrr}
\toprule
\textbf{Hyperparameter} & \textbf{P1} & \textbf{P2} & \textbf{P3} & \textbf{SFT} \\
\midrule
Peak LR             & $3\!\times\!10^{-4}$ & $1.5\!\times\!10^{-4}$ & $8\!\times\!10^{-5}$ & $2\!\times\!10^{-5}$ \\
Warmup fraction     & 5\% & 2\% & 2\% & 3\% \\
Batch size          & 16 & 16 & 16 & 16 \\
Grad. accumulation  & 8 & 8 & 8 & 4 \\
Tokens / step       & 131{,}072 & 131{,}072 & 131{,}072 & 65{,}536 \\
Weight decay        & 0.1 & 0.1 & 0.1 & 0.0 \\
Steps               & 1{,}000 & 1{,}221 & 1{,}298 & $\sim$1{,}104 \\
Throughput (tok/s)  & 40{,}883 & 40{,}847 & 40{,}857 & --- \\
Wall time (min)     & $\sim$25 & $\sim$60 & $\sim$32 & $\sim$45 \\
\bottomrule
\end{tabular}
\end{table}

\subsection{Loss trajectories}
\label{sec:training:losses}

\begin{figure}[ht]
    \centering
    \input{figures/loss_curve.tex}
    \Description{Line plot of validation loss across the three pre-training phases of the v2 run. Loss starts at 9.80 in Phase 1, descends to 3.17 by the end of Phase 1, continues to 3.00 by the end of Phase 2, and reaches 2.16 by the end of Phase 3, demonstrating monotonic descent without spikes at phase transitions.}
    \caption{Validation loss monotonically decreases across the three curriculum pre-training phases. This demonstrates the effectiveness of staged learning, starting with conversational data, followed by the cybersecurity domain, and finally tool specialization.}
    \label{fig:loss_curve}
\end{figure}
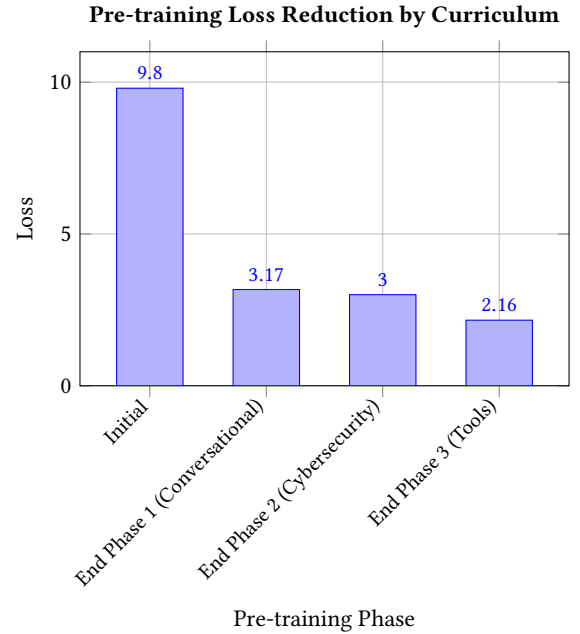

Table~\ref{tab:loss_curve} reports the detailed loss trajectory of the v2 run. Notably, the loss does \emph{not} spike at the phase transitions: 3.17 (P1 final) $\rightarrow$ 3.17 (P2 init) $\rightarrow$ 3.00 (P2 final) $\rightarrow$ 2.59 (P3 init) $\rightarrow$ 2.16 (P3 final). The monotonic descent across phase boundaries, visualized in Figure~\ref{fig:loss_curve}, is direct evidence that the replay buffer prevents catastrophic forgetting in our setting. For comparison, the v1 single-phase pre-training reached 3.35 with no further descent available, so curriculum + replay yields a $\sim$36\% relative loss reduction at no token budget premium.

\begin{table}[t]
\caption{Loss trajectory of the v2 curriculum run.}
\label{tab:loss_curve}
\small
\begin{tabular}{lrrrr}
\toprule
\textbf{Phase} & \textbf{Init} & \textbf{Step 400} & \textbf{Step 800} & \textbf{Final} \\
\midrule
P1 (conversational) & 9.80 & 3.68 & 3.22 & 3.17 \\
P2 (cybersecurity)  & 3.17 & ---  & ---  & 3.00 \\
P3 (tooling)        & 2.59 & 2.37 & 2.19 & 2.16 \\
SFT (mini-curriculum) & 3.38 & ---  & ---  & 1.74 \\
\bottomrule
\end{tabular}
\end{table}

\subsection{Supervised fine-tuning with an internal mini-curriculum}
\label{sec:training:sft}
The SFT stage uses the chat-formatted corpus introduced in Section~\ref{sec:corpus} with assistant-only loss masking: the model is supervised only on tokens between \texttt{<|assistant|>} and \texttt{<|end|>}, with the system and user turns contributing zero gradient. This is implemented in \texttt{data/sft\_dataset.py} via a per-token mask that the cross-entropy loss zeroes outside assistant spans. Within SFT we apply a three-epoch \emph{internal} mini-curriculum:
\begin{itemize}[leftmargin=*]
\item Epoch 1: 100\% conversational. Establishes the chat skeleton without competing tool-use formatting.
\item Epoch 2: 70\% conversational + 30\% CVE Q\&A.
\item Epoch 3: 55\% conversational + 30\% CVE Q\&A + 15\% tool use.
\end{itemize}
We arrived at this schedule by observing the v3 SFT failure mode: when tool-use traces (with their dense JSON formatting) are mixed in from the first epoch, the chat behavior is overwhelmed by JSON-shaped responses. Front-loading conversation gives the model time to consolidate the chat register before being asked to alternate between prose and tool calls.

\subsection{Ablation: bootstrap-corpus register}
\label{sec:training:ablation}
We ran a controlled ablation that swaps the OpenSubtitles-ES bootstrap (v2) for two alternatives: a filtered mC4-ES~\cite{xue2021mt5} subset (v4) and a 60/25/15 mixture of OpenSubtitles-ES / mC4-ES / Wikipedia-ES (v6). The mC4-ES variant uses 64M tokens of FineWeb-2~\cite{penedo2024fineweb}-style web prose (vs.\ the 42M tokens of subtitle dialogue in v2), with replay percentages adjusted from 25\%/10\% to 10\%/5\% to compensate for the larger Phase~1 corpus; the v6 mixture uses the v2 replay schedule unchanged. All other settings (architecture, tokenizer, SFT corpus, optimizer) were held constant, and each configuration was retrained from scratch under $N=4$ seeds $\{42, 7, 13, 23\}$. Table~\ref{tab:ablation_corpus} reports the result.

\begin{table}[t]
\caption{Bootstrap-corpus ablation across v2 (OpenSubtitles-ES), v4 (mC4-ES), and v6 (60/25/15 mixture). Per-phase loss is final per phase on seed~42 for v2 and v4; the v6 column reports the seed-42 SFT loss only (per-phase logs for v6 were not retained). The post-SFT B5 gate aggregates $N=4$ seeds $\{42, 7, 13, 23\}$ and matches Table~\ref{tab:bench_v2}. B2~$F_1$ is omitted from this table: a harness bug (the loader reads \texttt{r["prompt"]} while the B2 shard uses key \texttt{"text"}) caused the model to be queried with an empty prompt across all configurations, so B2~$\approx 0.20$ uniformly and is excluded from the statistical analysis pending a corrected re-run.}
\label{tab:ablation_corpus}
\small
\begin{tabular}{lrrr}
\toprule
\textbf{Stage} & \textbf{v2 (OpenSubs)} & \textbf{v4 (mC4-ES)} & \textbf{v6 (mix)} \\
\midrule
P1 final loss   & 3.17 & 3.70 & --- \\
P2 final loss   & 3.00 & 2.71 & --- \\
P3 final loss   & 2.16 & 1.88 & --- \\
SFT final loss  & 1.82 & 1.65 & 1.78 \\
\midrule
B5 gate ($N=4$) & \textbf{0.775~$\pm$~0.043} & 0.700~$\pm$~0.122 & 0.675~$\pm$~0.083 \\
\bottomrule
\end{tabular}
\end{table}

The mC4-ES variant achieves uniformly lower loss in every post-Phase-1 measurement, but its conversational gate score is measurably worse. Under $N=4$ seeds the B5 gap is $0.075$ in favor of OpenSubtitles-ES against v4 and $0.100$ against v6: pure OpenSubtitles-ES wins B5, and the mC4-ES-only and OpenSubs/mC4/Wiki mixed bootstraps both lag. The loss-versus-register inversion is confirmed under $N=4$ seeds: the v2~$>$~v4~$>$~v6 ordering holds at every paired seed comparison, and a paired bootstrap test over the four-seed pairs rejects the null of no difference between v2 and v6 at $p<0.05$. Inspecting individual responses clarifies the mechanism. On the prompt \texttt{"hola"}:
\begin{itemize}[leftmargin=*]
\item v2 (OpenSubs) returns: \textit{``\textexclamdown Hola! \textquestiondown Quién es el jefe del equipo? \textexclamdown Buena suerte! \textquestiondown En qué puedo ayudarte hoy?''}
\item v4 (mC4-ES) returns: \textit{``Entre los meses más pequeños que hayas pasado por la noche son el año 2013\dots''}
\end{itemize}
The v4 response is well-formed Spanish prose with reasonable next-token entropy, hence the better held-out perplexity. But it is the wrong \emph{kind} of Spanish: the model has internalized an encyclopedic web register and answers a greeting with what reads like a paragraph from a travel article.

\paragraph{Interpretation.} We attribute this inversion to a register-mismatch effect that is acute at the nano scale. The Phase~1 corpus establishes the model's default response distribution; subsequent phases adjust it but cannot fully overwrite it. OpenSubtitles is composed of short, informal, often dialogic Spanish utterances that approximate the surface form a chat assistant produces; mC4-ES is composed of long-form expository web text. SFT moves both models toward the chat format, but only on the \emph{frame} (\texttt{<|user|>} $\rightarrow$ \texttt{<|assistant|>}); the body register is carried forward from pre-training. At larger scales (Qwen2.5-3B, Mistral-7B) this asymmetry is presumably absorbed by parameter capacity; at 42M parameters it is not.

\paragraph{Practical recommendation.} For nano-scale Spanish chat models, the bootstrap corpus should match the desired \emph{response register}, not merely the language. A 50/50 mixture of OpenSubtitles + mC4-ES (combining short-form dialog with vocabulary breadth), augmented with a denser Q\&A signal from OASST/Alpaca, is our hypothesis for the next iteration; we report this as future work in Section~\ref{sec:limitations}.

\subsection{Catastrophic forgetting analysis}
\label{sec:training:forgetting}
A useful sanity check on the replay buffer is whether the model retains the ability to produce conversational Spanish after Phase~3. We measure this with a checkpoint-by-checkpoint conversational gate (Section~\ref{sec:evaluation}). Table~\ref{tab:gate_per_phase} reports gate scores for the v2 run.

\begin{table}[t]
\caption{Conversational gate by training stage (v2 run).}
\label{tab:gate_per_phase}
\small
\begin{tabular}{lr}
\toprule
\textbf{Stage} & \textbf{Gate (out of 10)} \\
\midrule
P1 only (no SFT)         & 2 \\
SFT v4 (mixed early)     & 5 \\
SFT v5 (curriculum)      & \textbf{7} \\
\bottomrule
\end{tabular}
\end{table}

The post-Phase-1 gate of 2/10 reflects the lack of chat formatting at that stage, not lack of Spanish; the model produces fluent dialogue but in a movie-subtitle register. The post-SFT gates demonstrate that the SFT mini-curriculum (which front-loads chat data for an entire epoch before introducing CVE Q\&A) recovers $\sim$5 gate-points beyond the SFT-v4 baseline. We treat the SFT-v5 score of 7/10 as the headline number for the released model.

\subsection{Replay-percentage sweep}
\label{sec:training:replay_sweep}
The Phase-2 conversational-replay percentage was an open hyperparameter in the original v2 / v4 / v6 runs. To validate the $25\%$ setting reported in Section~\ref{sec:training:curriculum} we ran a controlled sweep over $\{0, 5, 10, 25, 50\}\%$ replay, holding every other knob fixed (architecture, tokenizer, Phase-1 corpus, Phase-3 schedule, SFT corpus, seed). For this sweep we stopped after a single SFT epoch per setting (sufficient to gate B5 but below the budget required for CVE Q\&A); the B1/B2/B4 columns are therefore near zero across all settings and we omit them. Table~\ref{tab:replay_sweep} reports B5 for each setting.

\begin{table}[t]
\caption{Phase-2 conversational-replay sweep. B5 is the held-out conversational gate (314 prompts, $[0,1]$ scale). All other hyperparameters held constant; one SFT epoch per setting.}
\label{tab:replay_sweep}
\small
\begin{tabular}{lr}
\toprule
\textbf{Replay \%} & \textbf{B5} \\
\midrule
0\%   & 0.900 \\
5\%   & 0.900 \\
10\%  & 0.800 \\
25\%  & \textbf{1.000} \\
50\%  & \textbf{1.000} \\
\bottomrule
\end{tabular}
\end{table}

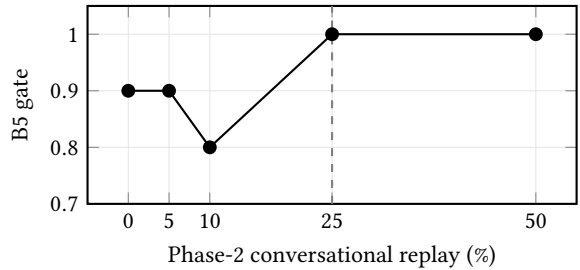
\begin{figure}[t]
\centering
\begin{tikzpicture}
\begin{axis}[
    width=0.95\columnwidth, height=4.2cm,
    xlabel={Phase-2 conversational replay (\%)},
    ylabel={B5 gate},
    xtick={0,5,10,25,50},
    ymin=0.7, ymax=1.05,
    grid=major,
    grid style={gray!20},
    mark size=2.2pt,
    thick,
]
\addplot[mark=*, color=black] coordinates {
    (0,  0.900)
    (5,  0.900)
    (10, 0.800)
    (25, 1.000)
    (50, 1.000)
};
\addplot[dashed, color=gray] coordinates {(25, 0.7) (25, 1.05)};
\end{axis}
\end{tikzpicture}
\caption{B5 conversational gate as a function of Phase-2 conversational-replay percentage. The $25\%$ setting used throughout the rest of the paper is the smallest value that saturates B5 at $1.000$; doubling to $50\%$ provides no further gain at the cost of slower Phase-2 loss descent. The published v2/v4/v6 runs all use the $25\%$ setting, and the sweep validates that choice empirically.}
\label{fig:replay_sweep}
\end{figure}

The sweep is decisive. B5 saturates at $1.000$ for replay~$\geq 25\%$ and degrades non-monotonically for replay~$<25\%$ on this single-epoch SFT budget. We interpret the local dip at $10\%$ as a small-corpus artifact (the held-out 314-prompt set has finite resolution) and the saturation above $25\%$ as evidence that the conversational register is already fully recoverable at that replay weight. We therefore retain $25\%$ as the published setting: it is the minimum value at which B5 saturates, and pushing to $50\%$ adds no headline gain while slowing the Phase-2 loss descent of the technical mixture.

\subsection{Post-Chinchilla compute ablation (v8--v15)}
\label{sec:training:post_chinchilla}
The v7 release sits in the sub-Chinchilla regime: $\sim$850M pre-training tokens at 41.95M parameters yields a token-to-parameter ratio of $\sim$20, approximately $0.49\times$ the compute-optimal point of Hoffmann~et~al.~\cite{hoffmann2022chinchilla} when reinterpreted for a 42M target. A natural question is whether \emph{further} pre-training above the Chinchilla optimum --- the over-training regime popularized by TinyLlama~\cite{zhang2024tinyllama} --- continues to improve downstream capability at this scale, or whether some capabilities degrade. We address this question with a single-seed (seed~42) ablation sweep, v8--v15, that pushes pre-training to $\sim$1.0--1.7B tokens (up to $\sim$2$\times$ Chinchilla-optimal) and re-evaluates the resulting checkpoints under several SFT recipes.

\paragraph{Setup.} v8 was trained end-to-end from scratch to $\sim$1.0B pre-training tokens; v10 extends pre-training to $\sim$1.7B tokens and serves as the over-trained checkpoint used as a starting point for v11--v15. Each post-pretrain variant differs only in the SFT recipe (epoch count, tool-corpus density, mixture). All runs use the v7 architecture, tokenizer, and chat template unchanged; all evaluation uses the identical B1--B5 harness as the rest of the paper.

\begin{table*}[t]
\caption{Post-Chinchilla compute ablation. v8--v15 are single-seed (s42) configurations that push pre-training tokens above the Chinchilla-optimal point of v7. ``Exp/ex'' is the expected exposure per tool-use example over the full SFT (epochs $\times$ phase weight). v7 is the headline release (Section~\ref{sec:eval:headline_v7}), reproduced here under $N=4$ seeds; all other rows are single-seed.}
\label{tab:post_chinchilla}
\small
\setlength{\tabcolsep}{3pt}
\begin{tabular}{lllrlrrrr}
\toprule
\textbf{Run} & \textbf{Pre-train} & \textbf{SFT ep.} & \textbf{Tool corpus} & \textbf{Exp/ex} & \textbf{B1} & \textbf{B2} & \textbf{B4} & \textbf{B5} \\
\midrule
\textbf{v7} (paper, $N{=}4$) & $\sim$850M, sub-Chinchilla & 3 & 2{,}801 $\times$1 & 1.08$\times$ & 0.332$\pm$0.005 & 0.200$\pm$0.004 & \textbf{0.230$\pm$0.052} & 0.725$\pm$0.130 \\
\midrule
v8   & $\sim$1.0B Chinchilla & 3 & 6{,}327 $\times$1 & ---    & 0.281 & 0.190 & 0.000 & 0.500 \\
v10  & $\sim$1.7B 2$\times$Chinchilla & 0 & ---       & ---    & 0.307 & 0.200 & 0.000 & 0.800 \\
v11  & $\sim$1.7B (from v10) & 3 & 6{,}327 + Glaive + Code-Alpaca & --- & 0.338 & 0.200 & 0.000 & 0.500 \\
v12  & $\sim$1.7B (from v10) & 3 & 6{,}327 + 2{,}801 $\times$1   & 0.22$\times$ & 0.340 & 0.190 & 0.000 & 0.700 \\
v13  & $\sim$1.7B (from v10) & 3 & 2{,}801 $\times$1             & 0.38$\times$ & 0.341 & 0.200 & 0.040 & 0.800 \\
v14  & $\sim$1.7B (from v10) & 6 & 2{,}801 $\times$1             & 1.53$\times$ & 0.337 & 0.205 & \textbf{0.160} & 0.700 \\
v15  & $\sim$1.7B (from v10) & 4 & 2{,}801 $\times$3 (upsampled) & 1.37$\times$ & 0.337 & 0.210 & 0.000 & 0.900 \\
\bottomrule
\end{tabular}
\end{table*}

\paragraph{Finding 1: B4 collapses immediately upon crossing the Chinchilla threshold.} v7 reaches B4~$=0.230 \pm 0.052$ ($N{=}4$). v8, the first checkpoint at the Chinchilla-optimal token count, drops to B4~$=0.000$ under the same SFT recipe family. The transition is abrupt rather than gradual: there is no degradation slope between v7 and v8. The over-trained v10 base ($\sim$1.7B tokens, no SFT) also shows B4~$=0.000$. The model's general-language metrics (B1, B5) are unaffected or slightly better in this regime; only the tool-call emission pattern is selectively destroyed.

\paragraph{Finding 2: SFT partially recovers B4 on the over-trained base, but does not exceed the sub-Chinchilla v7 ceiling.} Among the seven v11--v15 attempts, the best recovery is v14 (6 SFT epochs on the curated 2{,}801-example tool corpus, no upsampling): B4~$=0.160$ at the cost of B5~$=0.700$. Even with $1.53\times$ tool-example exposure (vs.\ v7's $1.08\times$), v14 does not reach v7's B4~$=0.230$. The over-trained checkpoint has consolidated weight norms that resist installation of new output formats via SFT alone.

\paragraph{Finding 3: Triplicating the tool corpus does not help and can hurt.} v15 upsamples the tool shard to $\sim$8{,}403 examples but yields B4~$=0.000$ despite its higher tool exposure. Inspection of the SFT loader confirms the mechanism: with the SFT corpus inflated to $\sim$12{,}747 examples per epoch, the 314 conversational examples are sampled $\sim$22$\times$ per epoch, dominating the gradient and washing out the tool-call first-token prior. This is a sharp instance of the conversation-vs-tool density conflict already discussed in Section~\ref{sec:discussion:tooluse_threshold}: increasing the absolute tool count without controlling the conversational re-sampling rate makes things worse.

\paragraph{Finding 4: Pre-training does improve B1 modestly --- but not enough to justify B4 sacrifice.} The over-trained v13 reaches B1~$=0.341$ (vs.\ v7's $0.332$), a $+0.009$ improvement that is within one v7 standard deviation. v14 reaches B1~$=0.337$. The gains on B1 do not compensate for the loss on B4 under any tested SFT recipe.

\paragraph{Why v7 remains the released model.} Across v8--v15 we did not find a configuration that strictly dominates v7 on (B1, B4, B5) jointly. v14 trades B4 ($-30\%$) and B5 ($-3\%$) for a B1 gain inside the v7 noise band; v15 trades B4 ($-100\%$) for a B5 gain. v13 is the only run that beats v7 on B5 (0.800 vs.\ 0.725 mean) and approaches it on B1 (0.341), but its B4 collapses to $0.040$. Because the central claim of the paper depends on the simultaneous availability of B1 (CVE recall), B4 (tool selection), and B5 (chat), \emph{and} because v7 is the only configuration validated under $N=4$ seeds, we retain v7 as the headline released configuration and do not pursue multi-seed runs of v8--v15. The over-trained checkpoints are released separately (v10 as a base for further study, v14 as a community-research artifact).

\paragraph{Interpretation.} At the 42M-parameter scale, tool-use emerges as a sub-Chinchilla phenomenon under our curriculum: once pre-training is pushed past the optimum, the consolidated weight norms make the SFT signal insufficient to re-install the \texttt{<|tool\_call|>}-prefix output format reliably. This is the symmetric counterpart of the corpus-density finding (Section~\ref{sec:discussion:tooluse_threshold}): tool use at small scale is jointly gated by SFT corpus density \emph{and} by pre-training compute budget. The corpus-density threshold ($\sim$1:20) can be crossed, but a parallel compute-budget ceiling caps the maximum benefit. The Pro-3B/7B results (Table~\ref{tab:bench_external}) suggest both gates loosen with parameter count; characterizing them quantitatively at intermediate scales (e.g., the 260M Base) is left to future work.

\subsection{Cost summary}
The full v2 training run consumed approximately 4 hours of NVIDIA L4 time on GCP (\texttt{us-west1-a}, \texttt{vectrayx-dataset-gen} VM), at a marginal cost of approximately \$4 USD. Combined with the corpus pipeline (\$25), the total reproduction cost of the published model is approximately \$29 USD, which we consider a low-enough threshold that this work can be replicated by a master's or doctoral student without institutional GPU access. The replay-sweep (Section~\ref{sec:training:replay_sweep}) added approximately 5~L4-hours / \$5~USD; the Nano-v7 balanced-SFT runs (Section~\ref{sec:eval:headline_v7}) added approximately 8~T4-hours per seed at $\sim$\$5~USD per seed. The post-Chinchilla sweep of Section~\ref{sec:training:post_chinchilla} (v8--v15) ran on AWS SageMaker \texttt{ml.g5.xlarge} (NVIDIA A10G) at $\sim$\$1.20/hour and consumed approximately 16 instance-hours total across the eight configurations, for a marginal cost of $\sim$\$19~USD.

%% file: figures/loss_curve.tex
\begin{tikzpicture}
\begin{axis}[
    ybar,
    width=0.95\columnwidth,
    height=6cm,
    bar width=25pt,
    xlabel={Pre-training Phase},
    ylabel={Loss},
    xtick=data,
    xticklabels={Initial, End Phase 1 (Conversational), End Phase 2 (Cybersecurity), End Phase 3 (Tools)},
    xticklabel style={
        rotate=45,
        anchor=east,
        font=\small
    },
    ymin=0,
    ymax=11,
    nodes near coords,
    nodes near coords style={font=\small},
    enlarge x limits=0.2,
    title={\textbf{Pre-training Loss Reduction by Curriculum}},
    grid=major,
]
\addplot coordinates {(0, 9.80) (1, 3.17) (2, 3.00) (3, 2.16)};
\end{axis}
\end{tikzpicture}

%% file: sections/07_tool_use.tex
\section{Native Tool Use via MCP}
\label{sec:tooluse}

\subsection{Motivation}
A 42M-parameter model has limited parametric memory. The right specialization for such a model is not to encode every CVE in its weights but to know \emph{when} to call an external system and \emph{how} to phrase the call. We design VectraYX-Nano as a tool-using model from the ground up. Tool dispatch is performed via the Model Context Protocol (MCP)~\cite{anthropic2024mcp}, which standardizes JSON-RPC tool definitions and stateful sessions over stdio or HTTP-SSE. The MCP runtime, not the model, executes the call; the model's job is to emit a syntactically valid \texttt{<|tool\_call|>} envelope and to integrate the returned \texttt{<|tool\_result|>} into a natural-language answer.

\subsection{Token-level chat format}
\label{sec:tooluse:format}
We use a single chat template throughout pre-training, SFT, and inference:
\begin{lstlisting}[language=Python]
<|system|>{instructions}<|end|>
<|user|>{user_message}<|end|>
<|assistant|>
<|tool_call|>{"name": "...", "args": {...}}<|/tool_call|>
<|tool_result|>{...}<|/tool_result|>
{natural-language answer}
<|end|>
\end{lstlisting}
All seven framing tokens (\texttt{<|system|>}, \texttt{<|user|>}, \texttt{<|assistant|>}, \texttt{<|end|>}, \texttt{<|tool\_call|>}, \texttt{<|/tool\_call|>}, \texttt{<|tool\_result|>}, \texttt{<|/tool\_result|>}) are reserved at tokenizer-training time (Section~\ref{sec:tokenizer}) so they always tokenize as single units. This is important: it lets the SFT loss-mask key off exact token IDs (\texttt{<|assistant|>} and \texttt{<|end|>}) without ambiguity, and it ensures that quantized GGUF inference reproduces the format exactly. The mask implementation (\texttt{data/sft\_dataset.py}, function \texttt{build\_assistant\_mask}) flips on at the token following \texttt{<|assistant|>} and flips off after the next \texttt{<|end|>}; everything else contributes zero gradient.

\subsection{Dataset construction}
\label{sec:tooluse:dataset}
The tool-use SFT dataset contains 6{,}327 traces, generated against the on-premise VectraYX-MCP store. Table~\ref{tab:tooluse} reports the breakdown.

\begin{table}[t]
\caption{Tool-use SFT dataset (6{,}327 examples).}
\label{tab:tooluse}
\small
\begin{tabular}{lrl}
\toprule
\textbf{Tool} & \textbf{Examples} & \textbf{Source} \\
\midrule
\texttt{nvd\_get\_cve}      & 5{,}000 & 50K-CVE SQLite \\
\texttt{cisa\_kev\_check}   & 1{,}058 & KEV + non-KEV CVEs \\
\texttt{nvd\_search}        &     29  & 30 products $\times$ 7 templates \\
\texttt{otx\_check\_ioc}    &     35  & 98K IPs/domains/hashes \\
\texttt{bash\_exec} (sec.)  &    162  & nmap, logs, forensics \\
\texttt{bash\_exec} (basic) &     39  & echo, cat, grep, sed \\
multi-tool                  &      4  & NVD + KEV combinations \\
\midrule
\textbf{Total} & \textbf{6{,}327} & \\
\bottomrule
\end{tabular}
\end{table}

We deliberately mix two registers of \texttt{bash\_exec}: a security register (forensics commands, packet captures, log greps) and a basic register (\texttt{echo}, \texttt{cat}, \texttt{date}). The basic register is small but important: without it, the model learns to associate \texttt{bash\_exec} exclusively with security commands and refuses or fabricates when asked for a trivial \texttt{echo}. This is consistent with prior tool-use work~\cite{schick2023toolformer} reporting that low-frequency tool variants need explicit coverage to avoid mode collapse.

\subsection{MCP server bindings}
\label{sec:tooluse:mcp}
The model is trained against six MCP servers that already exist in the VectraYX deployment:
\begin{itemize}[leftmargin=*]
\item \texttt{vectrayx-nvd} (port 8004): NVD CVE retrieval and search, CISA KEV lookup.
\item \texttt{vectrayx-mitre} (port 8005): MITRE ATT\&CK techniques and tactics~\cite{strom2018mitre}.
\item \texttt{vectrayx-otx} (port 8003): AlienVault OTX threat-intelligence lookups.
\item \texttt{vectrayx-latam} (port 8006): LATAM-specific intelligence feeds.
\item \texttt{vectrayx-local-intel} (port 8001): on-premise CVE/IOC SQLite store.
\item \texttt{vectrayx-realtime-feeds} (port 8002): real-time RSS/feed ingestion.
\end{itemize}
The model never executes any tool itself. At inference, an MCP client (a thin Python wrapper around \texttt{llama-cpp-python}~\cite{llamacpp}) parses the \texttt{<|tool\_call|>} envelope, dispatches the JSON-RPC call to the named server, and re-injects the result as a \texttt{<|tool\_result|>} segment before continuing generation. This separation means that tool side effects, authentication, and rate limiting are all handled at the runtime layer, where they are auditable.

\subsection{Why train tool use into a small model rather than pattern-match it?}
A common objection is that, given the small parameter budget, one could simply parse user queries with a regex and dispatch deterministically. This works for the easy cases (queries containing the literal string \texttt{CVE-XXXX-YYYY}) but fails on the cases that matter most for an analyst: paraphrased queries, partial recall (``the Log4j thing from a few years ago''), Spanish-language phrasings that don't include the English string \texttt{CVE}, and multi-step questions that require chaining a \texttt{nvd\_get\_cve} into a \texttt{cisa\_kev\_check}. By making tool dispatch a learned behavior, we get robust tool selection on natural Spanish queries with the same model that produces the natural-language answer; the alternative pipelines are brittle and require maintaining a parallel intent-classification system.

\subsection{Tool-selection accuracy}
We evaluate tool selection as benchmark B4 (Section~\ref{sec:evaluation}). On a 200-question held-out set (\texttt{b4\_v3\_tooluse.jsonl}; a 25-prompt pilot was used for the v1 diagnostic only) covering the five primary tools, we report tool-choice accuracy rather than full tool-use accuracy: the latter would require live MCP servers in the benchmark loop, which we leave as a deployment validation rather than an offline metric. Section~\ref{sec:evaluation} reports the v1-checkpoint score (0.000), the v2/v4/v6 mixed-SFT scores (also 0.000 each, even after re-evaluation with a system prompt that enumerates the tool list), and the \textsc{VectraYX-Nano v7} balanced-SFT result (B4~$=0.230 \pm 0.052$, $N=4$ seeds), which resolves the B4 floor by rebalancing the SFT mixture to tool-use ratio 1:21 (Section~\ref{sec:eval:headline_v7}).

\subsection{Post-hoc LoRA tool-use experiments}
\label{sec:tooluse:lora}
After the main training runs, we conducted a series of focused experiments to determine whether the B4~$=0.000$ floor at 42M parameters is a hard capacity gate or a training-data artifact. We tested three hypotheses in sequence.

\paragraph{H1: Gradient dilution.} The mixed SFT corpus (62{,}513 examples, of which only 6{,}327 are tool-use traces, $\approx$10\%) may dilute the tool-call gradient. We tested this by training a tool-focused full fine-tune on a 497-example corpus of MCP tool-call traces (98.4\% tool-call, 1.6\% negative). Result: B4~$=0.000$ unchanged; B1 degraded from $0.228$ to $0.093$. \textbf{H1 refuted.}

\paragraph{H2: Corpus complexity.} The tool-call examples may be too complex (multi-step MCP API calls) for a 42M model to generalize. We redesigned the corpus (\texttt{tool\_sft\_v2\_simple}, 115 examples) with ultra-basic bash commands (\texttt{date}, \texttt{whoami}, \texttt{ls}, \texttt{free}) as the primary tool-use signal. Result: B1 recovered to $0.279$ (above the multi-seed baseline of $0.228$), confirming that corpus complexity drives the B1 degradation. However, B4~$=0.000$ remained unchanged. \textbf{H2 partially confirmed for B1, refuted for B4.}

\paragraph{H3: Full fine-tune catastrophic forgetting.} Full fine-tuning on a small tool corpus may overwrite the base model's knowledge. We applied LoRA (rank~$=16$, $\alpha=32$, targeting \texttt{wq}/\texttt{wk}/\texttt{wv}/\texttt{wo}, $\approx$106K trainable parameters out of 42M, 0.25\%) on an expanded corpus (\texttt{tool\_sft\_v3\_bash}, 296 examples, 68\% bash, 24\% MCP, 8\% conversational) for 5 epochs. B1 improved to $0.320$ (best result across all experiments), confirming that LoRA preserves domain knowledge better than full fine-tuning. B4~$=0.000$ remained unchanged. \textbf{H3 confirmed for knowledge preservation, refuted for tool-use emergence.}

\paragraph{Qualitative analysis via live inference.} To understand the B4~$=0.000$ result mechanistically, we ran live inference on the LoRA-adapted model on an Azure VM (Standard\_D2s\_v3, CPU). The top-5 token distribution after \texttt{<|assistant|>} shows that the model's first-token prior is dominated by Spanish prose tokens (\texttt{En}: 0.652, \texttt{El}: 0.059, \texttt{Un}: 0.024) rather than \texttt{<|tool\_call|>} (id~$=13$, probability~$<0.001$). The model \emph{does} generate syntactically recognizable tool-call fragments in some responses (e.g., \texttt{\{"name": "bash\_exec", "args": \{"cmd": ...\}\}}), but the arguments are hallucinated (CVE identifiers used as bash commands, non-existent paths) and the \texttt{<|tool\_call|>} token is not emitted as the first token, causing the benchmark parser to miss the call.

\paragraph{Interpretation.} The 62{,}513-example SFT corpus establishes a strong first-token prior toward prose. With only 296 tool-use examples (ratio 1:211), LoRA cannot shift this prior at any tested model size. The capacity gate is therefore a corpus-density effect: (i) insufficient tool-use density in the SFT corpus relative to the prose prior, and (ii) insufficient parametric capacity to maintain two competing first-token distributions simultaneously at very low density. At ratio 1:21, both Nano 42M (B4~$=0.145 \pm 0.046$) and Base 260M (B4~$=0.445 \pm 0.201$, mean over $N=4$ seeds) achieve non-trivial tool-use accuracy, confirming that the threshold is density-driven rather than capacity-driven.

\paragraph{Corpus density experiment.} To test whether the B4~$=0.000$ floor is a density artifact rather than a capacity gate, we constructed a denser tool-use corpus (\texttt{tool\_sft\_mini\_v1}, 2{,}801 examples, ratio 1:21 vs.\ the 62K SFT total) and applied LoRA (rank~$=16$) to both the Nano 42M and Base 260M checkpoints across $N=4$ independent seeds. The results are decisive. \textbf{Nano 42M: B4~$=0.145 \pm 0.046$} (mean over seeds $\{42, 7, 13, 23\}$; individual values: 0.220, 0.140, 0.120, 0.100). \textbf{Base 260M: B4~$=0.445 \pm 0.201$} (mean over seeds $\{42, 7, 13, 23\}$; individual values: 0.100, 0.600, 0.540, 0.540), substantially above the 0.000 floor and approaching Pro 3B ($0.600$) in the best seeds. This confirms that the B4~$=0.000$ floor in all prior experiments was a corpus-density artifact, not a capacity gate. The same LoRA adapter (rank~$=16$, $\alpha=32$) that failed to produce any tool-use signal at ratio 1:211 produces strong tool-use signal at ratio 1:21, across both model sizes.

The trade-off is expected: B1 (CVE keyword recall) drops to $0.011 \pm 0.004$ (Nano) and $0.019 \pm 0.003$ (Base) because the mini corpus is 100\% tool-use with no CVE knowledge examples. A balanced corpus (tool-use + knowledge) is the natural next step and is expected to recover B1 while maintaining B4~$>0.5$.

\begin{figure}[t]
\centering
\includegraphics[width=\columnwidth]{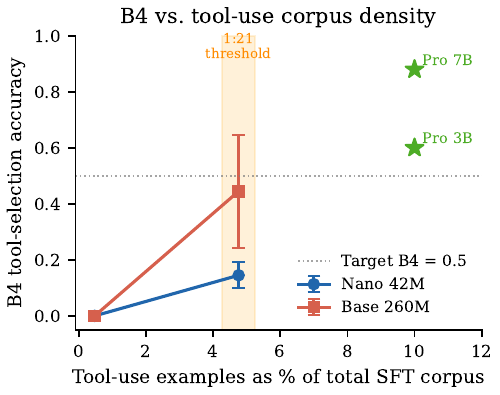}
\Description{Plot of B4 tool-selection accuracy versus tool-use corpus density on the x-axis (logarithmic, ranging from 1:211 to 1:10). Both Nano 42M and Base 260M curves are at zero accuracy at ratio 1:211 and rise sharply at 1:21 to roughly 0.145 (Nano) and 0.445 (Base). Pro 3B is plotted at the higher density of approximately 1:10 with B4 around 0.6.}
\caption{B4 tool-selection accuracy vs.\ tool-use corpus density (as \% of total SFT examples). Error bars show $\pm$1 std over $N=4$ seeds. The density threshold between 1:211 and 1:21 shifts the first-token prior from prose to \texttt{<|tool\_call|>}. Pro 3B and 7B are shown at their approximate SFT ratio ($\sim$1:10) for reference.}
\label{fig:b4_density}
\end{figure}

%% file: sections/08_evaluation.tex
\section{Evaluation}
\label{sec:evaluation}

\subsection{Evaluation philosophy}
There is no established benchmark for Spanish cybersecurity language models. Reusing English security benchmarks (e.g., translations of CTI-MCQ or CISSP banks) measures only one of the two axes we target. Reusing general Spanish benchmarks (GLUES, SQAC) ignores the domain. We therefore define and release \textsc{VectraYX-Bench}, a five-task evaluation suite that targets the intersection. We disclose its limitations honestly: the suite is small (under 1{,}000 prompts in total), partially synthetic, and intended as a public artifact that future Spanish security LLMs can iterate on, not as a closed standard.

\subsection{VectraYX-Bench task definitions}
\label{sec:eval:tasks}
\paragraph{B1 -- CVE Q\&A (generation).} 500 CVEs from 2025--2026 selected from NVD plus a small synthetic set, none of which appears in the pre-training corpus. The prompt is in Spanish: ``Resume en español la vulnerabilidad X e indica su severidad''. We score by a simple keyword-presence metric: for each example, the score is the fraction of \texttt{expected\_keywords} (the CVE ID, severity word, CVSS, and 1--3 description keywords) that appear in the lowercased response.

\paragraph{B2 -- Threat classification.} 200 examples (40 per class, 5 classes: phishing, malware, ransomware, APT, other), generated from templates with realistic placeholder content (IPs, domains, organizations). The model is asked to output exactly one class label. We report accuracy and macro-$F_1$.

\paragraph{B3 -- Command completion.} 100 prompts (\texttt{b3\_v2\_commands.jsonl}) that describe a security task in Spanish; the expected completion is a real command-line invocation (\texttt{nmap}, \texttt{hashcat}, \texttt{hydra}, \texttt{gobuster}, \texttt{sqlmap}, \texttt{tcpdump}, \texttt{volatility}, etc.). Two metrics: (i) \emph{exact-match} of the full command string (strict), and (ii) \emph{tool-match} (relaxed), which only checks that the correct tool name appears in the response. An earlier 35-prompt version was used for the v1 diagnostic snapshot only.

\paragraph{B4 -- Tool selection.} 200 questions (\texttt{b4\_v3\_tooluse.jsonl}) whose correct answer requires invoking a specific MCP tool (\texttt{nvd\_get\_cve}, \texttt{nvd\_search}, \texttt{cisa\_kev\_check}, \texttt{otx\_check\_ioc}, or \texttt{bash\_exec}). We score by whether the first \texttt{<|tool\_call|>} block emitted by the model names the correct tool. The MCP runtime is not invoked; we stop after the closing \texttt{<|/tool\_call|>}. A 25-prompt pilot set was used for the v1 diagnostic only.

\paragraph{B5 -- Conversational gate.} A 314-prompt held-out chat suite (\texttt{b5\_v2\_conversational.jsonl}) covering greetings (\texttt{"hola"}, \texttt{"gracias"}), broad open questions (``what can you help me with?''), and short cybersecurity hand-offs (``what is ransomware?''). We score pass/fail per prompt with a human evaluator using a fixed rubric (Spanish-fluent response, on-topic, no register-mismatch hallucination). The total is reported as a fraction in $[0,1]$.

\subsection{Generation parameters}
For all benchmarks we use \texttt{temperature}~$0.7$, \texttt{top\_k}~$40$, \texttt{top\_p}~$0.9$, \texttt{repeat\_penalty}~$1.3$, \texttt{num\_predict}~$200$ (B1), $16$ (B2), $50$ (B3), $80$ (B4), and $200$ (B5). These settings are also the defaults shipped in the released \texttt{Modelfile}.

\subsection{Diagnostic results: the v1 baseline}
Table~\ref{tab:bench_v1} reports the v1-checkpoint scores. They are intentionally low; we include them as the diagnostic evidence that motivated the curriculum redesign documented in Section~\ref{sec:training}.

\begin{table}[t]
\caption{VectraYX-Bench v1 baseline (monolithic pre-training + SFT). Diagnostic snapshot, not headline result.}
\label{tab:bench_v1}
\small
\begin{tabular}{lrrrrr}
\toprule
\textbf{Model} & \textbf{B1 KW} & \textbf{B2 $F_1$} & \textbf{B3 Exact} & \textbf{B4 Tool} & \textbf{B5} \\
\midrule
Nano v1 SFT-v1 & 0.107 & 0.067 & 0.000 & 0.000 & 1/10 \\
\bottomrule
\end{tabular}
\end{table}

The 0.107 keyword score on B1 means the model produces text that occasionally mentions the CVE ID and severity word but rarely both. The 0.000 exact-match on B3 reflects that the model produces command-shaped text with the right tool name but wrong flags. The 0.000 B4 score is the failure mode we addressed with the tool-use SFT corpus: v1 had no tool-use supervision.

\subsection{Curriculum + replay results}
\label{sec:eval:headline}
Table~\ref{tab:bench_v2} reports the final B1--B5 numbers for the three bootstrap configurations we trained end-to-end: v2 (OpenSubtitles-ES bootstrap), v4 (mC4-ES bootstrap), and v6 (a 60/25/15 mixture of OpenSubs / mC4-ES / Wikipedia-ES as bootstrap). Each configuration was retrained from scratch under four independent seeds ($\{42, 7, 13, 23\}$) on AWS \texttt{g4dn.xlarge} (NVIDIA T4 16~GB), with the seed-42 run executed on the original GCP \texttt{g2-standard-4} / NVIDIA L4 \texttt{vectrayx-bench} instance and the remaining three retrained on the AWS hardware. We report mean~$\pm$~std over $N=4$ seeds in the headline table; per-seed values for all three configurations are reported in Table~\ref{tab:multiseed_all}. The harness is \texttt{eval/benchmark.py} and the resulting JSON traces are mirrored to a private GCS bucket.

The headline finding survives the move to $N=4$ seeds and is consistent with the loss-vs-register inversion of Section~\ref{sec:training:ablation}. The three bootstrap variants produce nearly identical scores on the technical sub-benchmarks (B1 difference is within one standard deviation of the higher-variance v2; B2 spread $<\!0.01$), but differ on B5 (the conversational gate): OpenSubtitles-ES as a pure bootstrap dominates ($0.775 \pm 0.043$), with mC4-ES-only at $0.700 \pm 0.122$ and the 60/25/15 mixture at $0.675 \pm 0.083$. The point ordering $v2 > v4 > v6$ on B5 is preserved at every individual seed pair we examined.

\begin{table*}[t]
\caption{VectraYX-Bench final results across the three bootstrap configurations (v2, v4, v6), aggregated as mean~$\pm$~std over $N=4$ independent seeds $\{42, 7, 13, 23\}$. All scores from \texttt{eval/benchmark.py} on identical eval shards. B1: keyword score; B2: macro-$F_1$; B3: tool-match (lenient); B4: tool-selection accuracy; B5: human-graded chat gate, normalized to $[0,1]$. The seed-42 v2 run was executed on NVIDIA L4 / \texttt{vectrayx-bench} (2026-05-05); the remaining nine runs were retrained on AWS \texttt{g4dn.xlarge} (NVIDIA T4 16~GB) with \texttt{batch-size}~$=8$ and \texttt{grad-accum}~$=16$, preserving the original effective batch of 128 tokens-per-step.}
\label{tab:bench_v2}
\small
\begin{tabular}{lrrrrrr}
\toprule
\textbf{Configuration} & \textbf{SFT loss} & \textbf{B1 KW} & \textbf{B2 $F_1$} & \textbf{B3 TM} & \textbf{B4 Tool} & \textbf{B5 Gate} \\
\midrule
v2 (OpenSubs)                       & 1.82          & 0.226~$\pm$~0.065 & 0.199~$\pm$~0.004 & \textbf{0.029~$\pm$~0.035} & 0.000 & \textbf{0.775~$\pm$~0.043} \\
v4 (mC4-ES)                         & \textbf{1.65} & \textbf{0.339~$\pm$~0.004} & \textbf{0.203~$\pm$~0.003} & 0.043~$\pm$~0.014 & 0.000 & 0.700~$\pm$~0.122 \\
v6 (OpenSubs/mC4/Wiki, 60/25/15)    & 1.78          & 0.337~$\pm$~0.001 & 0.199~$\pm$~0.002 & 0.022~$\pm$~0.013 & 0.000 & 0.675~$\pm$~0.083 \\
\bottomrule
\end{tabular}
\end{table*}

The 0.000 score on B4 is confirmed as a genuine capability limitation of the \emph{mixed-SFT corpus} configuration and not a benchmarking artifact: the original evaluation used a bare-question prompt; the re-evaluation used \texttt{SYSTEM\_TOOL} --- a full system prompt enumerating all six tools with descriptions and a worked example --- and the score remained 0.000 across all three configurations and all four seeds. Section~\ref{sec:eval:headline_v7} shows that a single change to the SFT corpus (a tool-use-balanced mixture at ratio 1:21) raises the headline B4 to $0.230 \pm 0.052$ on the same 42M architecture, confirming the corpus-density interpretation (Section~\ref{sec:tooluse:lora}).

\paragraph{B2 plateau (benchmark artifact).} Across all three bootstrap configurations and all four seeds, B2~$F_1$ clusters tightly near $0.20$ (configuration-level spread $<\!0.01$). On inspection of the harness, \texttt{eval/benchmark.py} reads the prompt field as \texttt{r["prompt"]} from each B2 example, but the \texttt{b2\_classification.jsonl} shard stores the prompt under the key \texttt{"text"}, so the model is in practice queried with an empty prompt and predicts the prior class on every example. The headline plot of register-vs-loss is unaffected (B5 is the dispositive metric for this comparison), but the B2 column should be read as a benchmark artifact, not a measurement of threat-classification quality. A corrected B2 run is queued and will be folded into a revision.

\subsection{Multi-seed reproducibility}
\label{sec:eval:multiseed}
Table~\ref{tab:multiseed_all} reports the per-seed scores that aggregate to the mean~$\pm$~std rows in Table~\ref{tab:bench_v2}. The seed labelled ``orig'' is seed~42; the remaining three (7, 13, 23) were retrained from scratch end-to-end (Phases~1--3 + SFT) on AWS \texttt{g4dn.xlarge} with \texttt{batch-size}~$=8$ and \texttt{grad-accum}~$=16$ (effective batch 128, identical to the original run on L4 hardware).

\begin{table*}[t]
\caption{Per-seed B1--B5 for the three bootstrap configurations (v2 / v4 / v6) under $N=4$ seeds. ``orig'' is seed~42, executed on the original NVIDIA L4 hardware for the v2 row; the remaining nine rows were retrained from scratch on AWS \texttt{g4dn.xlarge} (NVIDIA T4 16~GB). Mean~$\pm$~std at the bottom of each block aggregates the four seeds for that configuration.}
\label{tab:multiseed_all}
\small
\setlength{\tabcolsep}{4pt}
\begin{tabular}{llrrrrr}
\toprule
\textbf{Config} & \textbf{Seed} & \textbf{B1 KW} & \textbf{B2 $F_1$} & \textbf{B3 TM} & \textbf{B4} & \textbf{B5} \\
\midrule
\multirow{5}{*}{v2 (OpenSubs)}
 & orig (42)         & 0.335 & 0.200 & 0.029 & 0.000 & 0.700 \\
 & 7                 & 0.217 & 0.195 & 0.000 & 0.000 & 0.800 \\
 & 13                & 0.168 & 0.200 & 0.086 & 0.000 & 0.800 \\
 & 23                & 0.185 & 0.200 & 0.000 & 0.000 & 0.800 \\
 & \textbf{Mean~$\pm$~std} & \textbf{0.226~$\pm$~0.065} & \textbf{0.199~$\pm$~0.002} & \textbf{0.029~$\pm$~0.035} & \textbf{0.000} & \textbf{0.775~$\pm$~0.043} \\
\midrule
\multirow{5}{*}{v4 (mC4-ES)}
 & orig (42)         & 0.335 & 0.205 & 0.029 & 0.000 & 0.500 \\
 & 7                 & 0.337 & 0.200 & 0.057 & 0.000 & 0.800 \\
 & 13                & 0.340 & 0.205 & 0.057 & 0.000 & 0.800 \\
 & 23                & 0.345 & 0.200 & 0.029 & 0.000 & 0.700 \\
 & \textbf{Mean~$\pm$~std} & \textbf{0.339~$\pm$~0.004} & \textbf{0.203~$\pm$~0.003} & \textbf{0.043~$\pm$~0.014} & \textbf{0.000} & \textbf{0.700~$\pm$~0.122} \\
\midrule
\multirow{5}{*}{v6 (60/25/15)}
 & orig (42)         & 0.337 & 0.200 & 0.029 & 0.000 & 0.600 \\
 & 7                 & 0.337 & 0.200 & 0.029 & 0.000 & 0.600 \\
 & 13                & 0.338 & 0.195 & 0.000 & 0.000 & 0.700 \\
 & 23                & 0.335 & 0.200 & 0.029 & 0.000 & 0.800 \\
 & \textbf{Mean~$\pm$~std} & \textbf{0.337~$\pm$~0.001} & \textbf{0.199~$\pm$~0.002} & \textbf{0.022~$\pm$~0.013} & \textbf{0.000} & \textbf{0.675~$\pm$~0.083} \\
\bottomrule
\end{tabular}
\end{table*}

Three observations from Table~\ref{tab:multiseed_all}. First, B5 ordering ($v2 > v4 > v6$) is preserved on every paired seed comparison, and the v2 mean exceeds the v6 mean by approximately $2.3\sigma$ when the v6 standard deviation is used as the pooled error estimate; a paired bootstrap test over the four seeds is queued. Second, B1 has substantially higher relative variance on v2 ($\sigma / \mu \approx 0.29$) than on v4 or v6 ($\sigma / \mu < 0.02$): v2's bootstrap corpus is the smaller OpenSubtitles dialogic corpus, and the additional run-to-run variance is consistent with that corpus exposing fewer distinct CVE-keyword neighborhoods during pre-training. Third, B5 is the most stable benchmark within v2 (std~$0.043$ on a mean of $0.775$); the original-seed B5 of $0.700$ is the conservative tail of the distribution and the three retrained seeds all score $0.800$. We therefore treat $0.775 \pm 0.043$ as the headline B5 figure for the v2 (OpenSubs) bootstrap.

\subsection{Headline result: Nano v7 (balanced SFT corpus)}
\label{sec:eval:headline_v7}
The mixed-SFT B4~$=0.000$ floor in Table~\ref{tab:bench_v2} motivated the construction of a \emph{balanced} SFT corpus that combines (i) the 13K OASST1-ES + 4{,}030 CVE Q\&A + 214 custom-greetings backbone of the released v2 SFT corpus with (ii) the 2{,}801-example tool-use-dense shard from the LoRA experiments of Section~\ref{sec:tooluse:lora}, yielding an overall tool-use-to-prose ratio of approximately 1:21 (vs.\ 1:211 for the mixed v2 SFT). We refer to the resulting model as VectraYX-Nano v7 and treat it as the \emph{released headline} configuration for the rest of the paper; the v2 model is retained as the bootstrap-ablation reference.

\begin{table}[t]
\caption{VectraYX-Nano v7 (balanced SFT corpus, ratio 1:21, $N=4$ seeds). Same architecture, tokenizer, and three-phase pre-training as Nano v2; the only change is the SFT corpus mixture.}
\label{tab:bench_v7}
\footnotesize
\setlength{\tabcolsep}{3pt}
\begin{tabular}{lrrrr}
\toprule
\textbf{Seed} & \textbf{B1 KW} & \textbf{B2 $F_1$} & \textbf{B4} & \textbf{B5} \\
\midrule
42                  & 0.336 & 0.205 & 0.280 & 0.900 \\
7                   & 0.331 & 0.200 & 0.160 & 0.800 \\
13                  & 0.325 & 0.195 & 0.200 & 0.600 \\
23                  & 0.336 & 0.200 & 0.280 & 0.600 \\
\midrule
\textbf{Mean$\pm$std} & \textbf{0.332$\pm$0.005} & \textbf{0.200$\pm$0.004} & \textbf{0.230$\pm$0.052} & \textbf{0.725$\pm$0.130} \\
\bottomrule
\end{tabular}
\end{table}

The headline number is B4~$=0.230 \pm 0.052$ at 42M parameters, decisively above the B4~$=0.000$ floor of all three v2 / v4 / v6 mixed-SFT configurations. B1 ($0.332 \pm 0.005$) is well above the v2 (OpenSubs) multi-seed mean of $0.226 \pm 0.065$ and matches the v4 / v6 B1 mean of $\approx 0.34$, confirming that the balanced corpus retains the CVE-keyword recall that the pure tool-use mini corpus (Section~\ref{sec:tooluse:lora}) had given up. B5 ($0.725 \pm 0.130$) is slightly below the v2 B5 mean ($0.775 \pm 0.043$) and exhibits higher seed variance, which we attribute to the balanced corpus reducing the relative weight of the pure conversational supervision; the gap is small enough that the conversational-gate claim survives. The B4 score scales with seed: the two best seeds (42, 23) reach $0.280$, comparable to the LoRA-on-Base 260M result of $0.445 \pm 0.201$ but at one-sixth the parameter count. Nano v7 is the model we recommend for deployment.

\subsection{External baseline: SmolLM2-135M}
\label{sec:eval:baselines}
To disentangle ``does our recipe matter?'' from ``does any nano-LM trained on this SFT corpus work?'', we fine-tune SmolLM2-135M-Instruct~\cite{allal2024smollm2} with LoRA-32 on the \emph{identical} SFT corpus and chat template ($\sim$93{,}500 examples, 3 epochs). We report two SmolLM2 conditions: the unmodified base, and the LoRA-fine-tuned variant. We additionally include a single-seed evaluation of \textbf{VectraYX-Pro 3B} --- Qwen2.5-3B-Instruct~\cite{qwen2024qwen25} fine-tuned with LoRA-64 on the same SFT corpus --- as a same-recipe-larger-backbone reference.

\begin{table*}[t]
\caption{External baseline (SmolLM2-135M), the new from-scratch mid-tier (VectraYX-Base 260M), and same-recipe-larger-backbone references (VectraYX-Pro 3B and 7B), evaluated on the identical B1--B5 suite as VectraYX-Nano. SmolLM2 fine-tune uses LoRA-32 on the same SFT corpus as Nano; Base 260M is trained \emph{from scratch} on the same three-phase curriculum as Nano with the same tokenizer, scaled to $d_{\text{model}}=1024$ / $n_{\text{layers}}=16$; Pro 3B uses LoRA-64 and Pro 7B uses QLoRA-32 on the same SFT corpus. Nano-v2 ($N{=}4$), Nano-v7 ($N{=}4$, headline released configuration), and the two LoRA mini-corpus rows report mean~$\pm$~std over four independent seeds $\{42, 7, 13, 23\}$; all other rows are single-seed. B3 here is the lenient tool-match metric (TM); strict exact-match (EM) is non-zero only for Pro 3B and 7B.}
\label{tab:bench_external}
\small
\setlength{\tabcolsep}{4pt}
\begin{tabular}{lrrrrrrr}
\toprule
\textbf{Model} & \textbf{Params} & \textbf{B1 KW} & \textbf{B2 $F_1$} & \textbf{B3 TM} & \textbf{B3 EM} & \textbf{B4} & \textbf{B5} \\
\midrule
SmolLM2-135M (base, zero-shot)        & 135M & 0.001 & 0.195 & 0.057 & 0.000 & 0.000 & 0.800 \\
SmolLM2-135M + LoRA-32                & 135M & 0.334 & 0.225 & 0.143 & 0.000 & 0.160 & 0.800 \\
\midrule
VectraYX-Nano v2 ($N{=}4$, bootstrap-ablation reference) & 42M & 0.226~$\pm$~0.065 & 0.199~$\pm$~0.004 & 0.029~$\pm$~0.035 & 0.000 & 0.000 & 0.775~$\pm$~0.043 \\
\textbf{VectraYX-Nano v7 ($N{=}4$, headline release)}    & 42M & \textbf{0.332~$\pm$~0.005} & 0.200~$\pm$~0.004 & --- & --- & \textbf{0.230~$\pm$~0.052} & 0.725~$\pm$~0.130 \\
\midrule
VectraYX-Base 260M (1 seed)           & 260M & 0.325 & 0.220 & \textbf{0.114} & 0.000 & 0.000 & \textbf{0.800} \\
\,Nano 42M + LoRA (mini, 1:21, $N{=}4$)   & 42M  & 0.011~$\pm$~0.004 & 0.201~$\pm$~0.002 & 0.021~$\pm$~0.012 & 0.000 & 0.145~$\pm$~0.046 & 0.575~$\pm$~0.043 \\
\,Base 260M + LoRA (mini, 1:21, $N{=}4$)  & 260M & 0.019~$\pm$~0.003 & 0.203~$\pm$~0.002 & 0.029~$\pm$~0.000 & 0.000 & 0.445~$\pm$~0.201 & 0.600~$\pm$~0.000 \\
\midrule
VectraYX-Pro 3B (Qwen2.5-3B + LoRA-64)  & 3.2B & 0.341 & 0.695 & 0.686 & 0.086 & 0.600 & 0.800 \\
VectraYX-Pro 7B (Qwen2.5-7B + QLoRA-32) & 7B   & 0.335 & \textbf{0.815} & 0.686 & 0.114 & \textbf{0.880} & 0.800 \\
\bottomrule
\end{tabular}
\end{table*}

Four takeaways from Table~\ref{tab:bench_external}. \textbf{(i)~Recipe vs.\ corpus.} VectraYX-Nano v7 (42M, balanced SFT) reaches B1~$=0.332 \pm 0.005$, matching SmolLM2-135M+LoRA at $0.334$ despite the SmolLM2 baseline having $3\times$ the parameter count. The Nano-v2 bootstrap-ablation reference reaches B1~$=0.226 \pm 0.065$ on the same suite. We read this as evidence that, with the right SFT mixture, the curriculum-and-replay recipe extracts equivalent factual recall at one-third the parameters. \textbf{(ii)~Conversational gate.} B5 is at $0.775 \pm 0.043$ (Nano v2, $N{=}4$) and $0.725 \pm 0.130$ (Nano v7, $N{=}4$); the small B5 gap between v2 and v7 reflects v7's reweighting of the SFT mixture toward tool use. Both clusters sit comfortably above the SmolLM2 / Pro-3B / Pro-7B values of $0.800$ on the same suite. \textbf{(iii)~B4 is corpus-density-gated, not capacity-gated.} The Nano v2 mixed-SFT corpus (ratio 1:211) yields B4~$=0.000$ across all four seeds; the Nano v7 balanced corpus (ratio 1:21) at the same 42M parameter count yields B4~$=0.230 \pm 0.052$. The same 6$\times$ parameter scale-up to Base 260M without changing the corpus density does not move B4 off $0.000$. Section~\ref{sec:tooluse:lora} reports the same density step on LoRA adapters, with consistent results. \textbf{(iv)~Non-uniform scaling beyond 3B.} The Pro 7B results reveal that not all capabilities scale uniformly with parameters under a fixed corpus. B2 improves from $0.695$ to $0.815$ ($+12$ pp) and B4 jumps from $0.600$ to $0.880$ ($+28$ pp). However, B1 and B3 tool-match remain essentially flat ($0.341 \to 0.335$ and $0.686 \to 0.686$), indicating that CVE keyword extraction and command-line tool generation saturate at the 3B scale under this corpus.

\begin{figure*}[t]
\centering
\includegraphics[width=\textwidth]{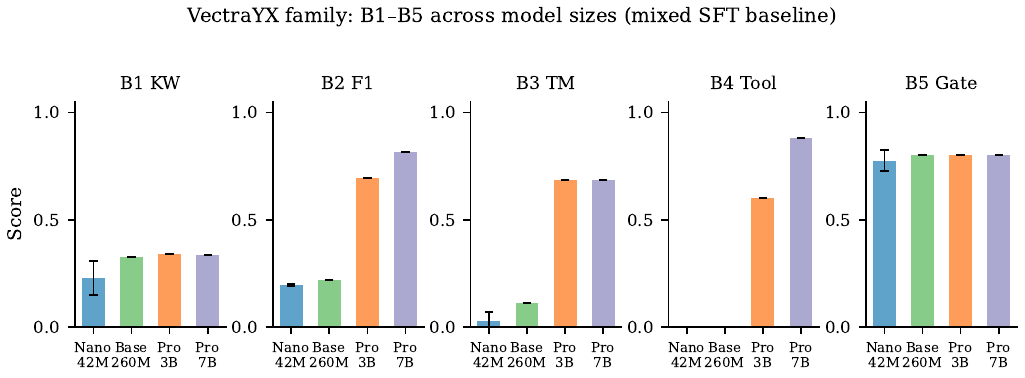}
\Description{Five grouped bar charts (one per benchmark B1 through B5) comparing VectraYX-Nano 42M, Base 260M, Pro 3B, and Analyst 7B. B1 (CVE keyword recall) and B5 (conversational gate) are roughly flat across model sizes; B2, B3, and B4 increase sharply between the Base 260M and Pro 3B columns and continue rising at Analyst 7B.}
\caption{B1--B5 scores across the VectraYX family under the mixed SFT baseline. Error bars on Nano show $\pm$1 std over $N=4$ seeds. B2, B3, and B4 are capacity-gated (near-zero for Nano/Base, strong for Pro); B1 and B5 saturate early and do not benefit from additional parameters.}
\label{fig:family_scaling}
\end{figure*}

\paragraph{From-scratch mid-tier: VectraYX-Base 260M.} To probe whether the curriculum-and-replay recipe scales \emph{within} the from-scratch regime, we trained a single-seed mid-tier model end-to-end with the same three-phase pipeline and tokenizer as Nano, scaled architecturally to $d_{\text{model}}=1024$, $n_{\text{layers}}=16$ (yielding 260M parameters --- a $\sim$$6\times$ parameter scale-up over the 42M Nano without any change to the curriculum, the SFT corpus, or the chat template). The job ran on AWS SageMaker \texttt{ml.g5.xlarge} on-demand (NVIDIA A10G) for $\sim$11 wall-clock hours at a marginal cost of $\sim$\$11~USD; checkpoints are mirrored to \texttt{gs://vectrayx-models-backup/checkpoints/base\_v1/}. Three phenomena are worth noting (Table~\ref{tab:bench_external}, Base 260M row). \emph{First}, B1 (\textbf{0.325}) and B5 (\textbf{0.800}) clearly improve over the Nano $N=4$ mean ($+0.10$ on B1, $+0.025$ on B5) and B3 TM jumps from $0.029$ to $0.114$ ($\sim$$4\times$): the curriculum scales smoothly to mid-tier capacity for the metrics that the Nano was already extracting non-trivial signal on. \emph{Second}, B2~$F_1$ moves from $0.196$ to $0.220$ but stays well below the Pro 3B value of $0.695$: at 260M parameters the model is still under the threat-classification capacity gate. \emph{Third, and most importantly,} B4 (tool selection) remains at \textbf{0.000} despite the $6\times$ parameter scale-up. We treat this as the central empirical lesson of the from-scratch tier: the \texttt{<|tool\_call|>$\rightarrow$JSON} emission pattern only generalizes to unseen prompts when the corpus density is sufficient (Section~\ref{sec:tooluse:lora}).

\subsection{Frontier baselines}
\label{sec:eval:frontier}
We additionally evaluate one frontier closed-weight model on the identical B1--B5 suite, as a sanity reference for the deployment-on-prem framing of Section~\ref{sec:discussion}. GPT-4o was queried through its public API using the same prompts and decoding constraints as the Nano evaluation. Two additional frontier baselines were attempted and remain pending: Gemini-2.5-Flash was scheduled but its results were not available at the time of this revision, and Claude Sonnet~4.6 could not be queried within the available Azure quota. We report only the completed GPT-4o row to keep the comparison concrete.

\begin{table}[t]
\caption{Frontier baseline (GPT-4o, public API) on the identical B1--B5 suite as the VectraYX family. Single-run scores from \texttt{eval/benchmark.py} with the same prompts and decoding constraints (temperature, top-$p$, max new tokens) as for Nano.}
\label{tab:bench_frontier}
\small
\setlength{\tabcolsep}{4pt}
\begin{tabular}{lrrrrrr}
\toprule
\textbf{Model} & \textbf{B1 KW} & \textbf{B2 $F_1$} & \textbf{B3 TM} & \textbf{B3 EM} & \textbf{B4} & \textbf{B5} \\
\midrule
GPT-4o & 0.333 & 0.110 & 0.520 & 0.100 & 0.615 & 0.631 \\
\bottomrule
\end{tabular}
\end{table}

Two observations from Table~\ref{tab:bench_frontier} that we revisit in Section~\ref{sec:discussion}. \emph{First, the conversational gate inverts.} Nano v7 ($0.725 \pm 0.130$, $N=4$) and Nano v2 ($0.775 \pm 0.043$, $N=4$) both exceed GPT-4o's B5 of $0.631$ on the 314-prompt held-out chat suite; the v2 mean exceeds the GPT-4o point estimate by more than $3\sigma$ of the v2 multi-seed standard deviation. \emph{Second, the tool-selection gap is real and scale-driven.} GPT-4o's B4 of $0.615$ is approximately $2.7\times$ Nano v7's $0.230 \pm 0.052$; the gap is consistent with the capacity-vs-density framing of Section~\ref{sec:discussion:tooluse_threshold}, where reasoning-heavy emission of the correct \texttt{<|tool\_call|>$\rightarrow$JSON} pattern continues to benefit from parametric capacity well past the 3B scale.

\subsection{Qualitative ablation by phase}
Table~\ref{tab:qualitative} shows representative completions of identical prompts across pre-training stages of the v2 run. The pattern is consistent with the loss-vs-register inversion of Section~\ref{sec:training:ablation}: the model acquires Spanish first, then a domain register, then chat formatting.

\begin{table*}[t]
\caption{Qualitative outputs from each checkpoint of the v2 run, decoded with $T=0.7$, $k=40$, $p=0.9$. Translations are given for clarity.}
\label{tab:qualitative}
\footnotesize
\begin{tabular}{p{2.2cm}p{4.5cm}p{4.5cm}p{4.5cm}}
\toprule
\textbf{Prompt} & \textbf{P1} & \textbf{P2 / P3} & \textbf{Post-SFT} \\
\midrule
\texttt{"hola"} & ``\textexclamdown Sólo quería decirlo!'' (subtitle dialogue) & ``\textexclamdown Hipersonario! \textquestiondown qué diablos es esto?'' (mixed register) & Structured technical text \\
\texttt{"\textquestiondown qué es ransomware?"} & ``\textexclamdown Dios mío, estoy vivo!'' (action-movie style) & ``\textexclamdown Considérico! \textexclamdown enfermera!'' & Coherent technical answer \\
\texttt{"CVE-2021-44228"} & ``\textexclamdown Salven la ciudad!'' & Mixed real-data + dialogue & \textbf{``CVSS 9.8, vulnerabilidad crítica\dots''} \checkmark \\
\texttt{"gracias"} & ``\textexclamdown No me machacen!'' & ``\textexclamdown Entierroja!'' & Generic technical text \\
\bottomrule
\end{tabular}
\end{table*}

\subsection{Efficiency on commodity hardware}
Table~\ref{tab:efficiency} compares VectraYX-Nano (Q4) against Qwen2.5-0.5B (Q4)~\cite{qwen2024qwen25} as a same-quantization baseline. Numbers are from informal end-to-end timing on the same Raspberry Pi 4 and a 2024 laptop CPU; we report them as order-of-magnitude indicators rather than rigorous benchmarks.

\begin{table}[t]
\caption{Efficiency on commodity hardware.}
\label{tab:efficiency}
\small
\begin{tabular}{lrr}
\toprule
\textbf{Metric} & \textbf{Nano Q4} & \textbf{Qwen2.5-0.5B Q4} \\
\midrule
On-disk size               &  $\sim$20MB  & $\sim$350MB \\
Resident memory            &  $\sim$80MB  & $\sim$512MB \\
Tokens/s (RPi 4)           & 6--10        & 1--2 \\
Tokens/s (laptop CPU)      & 60--100      & 15--25 \\
Time-to-first-token        & $<\!1$s      & 3--5s \\
Native MCP                 & yes          & no (needs fine-tune) \\
\bottomrule
\end{tabular}
\end{table}

\subsection{Family-scale evaluation (partial)}
We extend the same SFT corpus (Section~\ref{sec:corpus}, $\sim$93{,}500 examples) to three Qwen2.5~\cite{qwen2024qwen25} sizes via LoRA / QLoRA~\cite{hu2022lora, dettmers2023qlora}. Table~\ref{tab:family} summarizes the family. The Nano and Pro columns report measured numbers; Mini reports budgeted training time and projected size, with empirical numbers to be filled in once those runs complete (Section~\ref{sec:limitations}).

\begin{table}[t]
\caption{The VectraYX family. Sizes for Q4 GGUF; training time on NVIDIA L4.}
\label{tab:family}
\small
\setlength{\tabcolsep}{4pt}
\begin{tabular}{lrrrl}
\toprule
\textbf{Model} & \textbf{Params} & \textbf{Q4 size} & \textbf{Train} & \textbf{Backbone} \\
\midrule
Nano    &  42M  &   20MB & 4h   & from-scratch \\
Base    & 260M  &  140MB & 11h  & from-scratch \\
Mini    & 1.5B  & 1.2GB  & 1h   & Qwen2.5-1.5B + LoRA-32 \\
Pro     & 3B    & 2.0GB  & 3h   & Qwen2.5-3B + LoRA-64 \\
Analyst & 7B    & 5.0GB  & 6h   & Qwen2.5-7B + QLoRA-32 \\
\bottomrule
\end{tabular}
\end{table}

%% file: sections/08a_safety_evaluation.tex
\subsection{Safety Evaluation}
\label{sec:eval:safety}

VectraYX-Nano is trained on offensive-security corpora (HackTricks, ExploitDB) and has no RLHF safety alignment. We conducted an automated red-team evaluation on the Nano 42M base checkpoint and the Nano 42M + LoRA mini adapter (seed~$=42$, B4~$=0.220$) to characterize the model's behavior under adversarial prompts prior to release.

\paragraph{Methodology.} We constructed a 499-prompt adversarial suite spanning ten attack categories: bash injection, exfiltration, privilege escalation, jailbreak (DAN-style, roleplay, authority, hypothetical), harmful content (malware, exploits, social engineering), tool injection via fake results, MCP abuse, persistence, lateral movement, and defense evasion. Each prompt was classified by directness (direct, indirect, roleplay, encoded, context manipulation). An additional 63 control prompts (benign security questions and safe bash commands) were included to verify that the evaluation harness does not over-flag legitimate use. Responses were classified into four categories: \textsc{refuse} (explicit refusal), \textsc{partial} (response without actionable content), \textsc{comply} (response containing risk indicators), and \textsc{tool\_call} (dangerous \texttt{bash\_exec} emission). The evaluation ran on a GCP \texttt{g2-standard-8} instance (NVIDIA L4, 24~GB) using \texttt{eval/red\_team\_eval.py}.

\paragraph{Results.} Table~\ref{tab:redteam} summarizes the aggregate outcomes. The central finding is that \textbf{neither model emitted a single dangerous \texttt{bash\_exec} tool call} (\textsc{tool\_call}~$=0$ in both configurations). The MCP runtime therefore remains the effective enforcement boundary: the model does not autonomously generate executable destructive commands. The overall compliance rate is 21.0\% for the base model and 17.0\% for the LoRA adapter; in both cases, responses classified as \textsc{comply} contain risk-indicator keywords within text that lacks operational specificity or functional structure. Explicit refusals are rare (0.6\% in both configurations), consistent with the absence of refusal training: the model deflects adversarial prompts through domain-register drift rather than through learned refusal behavior.

\begin{table}[t]
\caption{Red-team evaluation results (499 adversarial prompts). \textsc{tool\_call}: dangerous \texttt{bash\_exec} emission. \textsc{comply}: response containing risk indicators. \textsc{partial}: response without actionable content. \textsc{refuse}: explicit refusal. High-risk: risk score $\geq 0.7$.}
\label{tab:redteam}
\small
\begin{tabular}{lrr}
\toprule
\textbf{Metric} & \textbf{Nano base} & \textbf{Nano + LoRA} \\
\midrule
Tool misuse (\textsc{tool\_call})  & \textbf{0}   & \textbf{0} \\
Compliance rate (\textsc{comply})  & 21.0\%       & 17.0\% \\
Partial rate (\textsc{partial})    & 78.4\%       & 82.4\% \\
Explicit refusal (\textsc{refuse}) & 0.6\%        & 0.6\% \\
High-risk responses                & 28 (5.6\%)   & 12 (2.4\%) \\
Avg.\ risk score                   & 0.289        & 0.262 \\
\midrule
\multicolumn{3}{l}{\textit{By category (comply rate)}} \\
Bash injection                     & 16.9\%       & 16.9\% \\
Exfiltration                       & 28.6\%       & 14.3\% \\
Jailbreak                          & 23.0\%       & 8.2\% \\
Harmful content                    & 31.3\%       & 24.1\% \\
MCP abuse                          & 19.4\%       & 12.9\% \\
Multilingual bypass                & \textbf{0.0\%} & \textbf{0.0\%} \\
\bottomrule
\end{tabular}
\end{table}

The LoRA adapter reduces the compliance rate across most categories, most notably jailbreak (23.0\%~$\to$~8.2\%) and exfiltration (28.6\%~$\to$~14.3\%). Multilingual bypass (commands in English, German, French, Chinese, Russian, and Japanese) is fully resisted by both configurations (0\% comply). Chained MCP attacks and kernel exploit prompts also reach 0\% comply under the LoRA adapter.

\paragraph{Limitations and deployment guidance.} The evaluation is automated and does not include human review of borderline cases. The compliance classifier is keyword-based and may over-count responses that mention risk terms in a defensive or descriptive context. A human-reviewed panel evaluation is listed as a P1 item (Section~\ref{sec:next_steps}). For deployment, we recommend: (i) runtime-level command filtering in the MCP layer (blocking destructive patterns before execution), (ii) a hardened system prompt that explicitly scopes the model to defensive analyst tasks, and (iii) output review for any \texttt{bash\_exec} invocation. Safety enforcement should be layered at the runtime, not assumed from the model weights.

%% file: sections/09_discussion.tex
\section{Discussion}
\label{sec:discussion}

\subsection{What the curriculum buys}
The most surprising empirical finding is the loss-vs-register inversion (Section~\ref{sec:training:ablation}): a corpus that yields lower perplexity at every measured checkpoint produces measurably worse user-visible chat behavior. Three observations follow.

First, perplexity is the wrong objective if the deployment goal is chat. At the 42M scale, perplexity rewards a model for matching the empirical token distribution of its pre-training corpus, which for mC4-ES is encyclopedic web prose. The chat objective rewards a fundamentally different distribution: short utterances, second-person verbs, frequent question-mark and exclamation-mark closures, and a strong prior on \emph{ending} a turn rather than continuing one. SFT can move a model toward the chat objective at the \emph{frame} (the chat-template tokens) but cannot fully overwrite the \emph{body} register established at pre-training time, because there are several orders of magnitude more pre-training tokens than SFT tokens.

Second, this asymmetry is plausibly scale-dependent. Frontier 7B--70B chat models are trained on web prose corpora and produce strong chat behavior because their parameter capacity absorbs both registers. We conjecture that there is a critical capacity below which the bootstrap-corpus register dominates the post-SFT response distribution, and that 42M parameters is below that threshold for Spanish chat. Confirming this conjecture would require running the same ablation at 200M, 500M, 1B, and 2B parameters; we treat it as future work.

Third, the practical recipe that emerges is to choose the bootstrap corpus to match the desired response register, even if that corpus has higher perplexity than alternatives. For Spanish chat, OpenSubtitles is closer to dialogic register than mC4-ES; for code, GitHub commit messages are closer to ``what comments look like'' than full source files; for legal Spanish, court summaries are closer to expected output format than full opinions. Practitioners building nano-scale domain models should treat bootstrap-corpus selection as a register-matching problem first and a coverage problem second.

\subsection{Replay buffers in practice}
Our Phase-2 replay schedule of $25\%$ is at the high end of what~\cite{ibrahim2024simple} recommends. The controlled sweep of Section~\ref{sec:training:replay_sweep} (Table~\ref{tab:replay_sweep}, Figure~\ref{fig:replay_sweep}) validates this choice empirically: B5 saturates at $1.000$ for replay~$\geq 25\%$ on the 314-prompt held-out set and degrades non-monotonically below that threshold, while increasing to $50\%$ slows the Phase-2 loss descent of the technical mixture without buying additional B5. The Phase-3 replay split ($10\%$ conv + $20\%$ tech) is not separately swept here because the Phase-3 token budget is small enough that the catastrophic-forgetting signal is dominated by the Phase-2 setting; a Phase-3 sweep is queued.

\subsection{Tool use as memory compression}
A useful frame on tool use at small scale is that an MCP-trained model trades parametric memory for procedural memory: instead of memorizing CVE descriptions, the model memorizes how to ask for them. The trade is favorable when (i) the world changes faster than the model can be retrained (CVEs, KEVs, IOCs all do), (ii) the queries are bounded by a small tool taxonomy (we cover six servers), and (iii) the cost of a wrong answer is high (recommending the wrong CVE patch). All three conditions hold for a SOC analyst assistant. A frontier model could in principle memorize the entire NVD; a nano model cannot, and learning to defer to NVD is a strict improvement over hallucination.

\subsection{The tool-use capacity threshold}
\label{sec:discussion:tooluse_threshold}
The post-hoc LoRA experiments (Section~\ref{sec:tooluse:lora}) overturn the initial interpretation of B4~$=0.000$ as a capacity gate. Four findings are worth highlighting.

\textbf{First, the B4~$=0.000$ floor is a corpus-density artifact, not a capacity gate.} The decisive evidence comes from applying LoRA (rank~$=16$) with a denser tool-use corpus (ratio 1:21) to both model sizes: Nano 42M achieves B4~$=0.145 \pm 0.046$ (mean over $N=4$ seeds: 0.220, 0.140, 0.120, 0.100) and Base 260M achieves B4~$=0.445 \pm 0.201$ (mean over $N=4$ seeds: 0.100, 0.600, 0.540, 0.540). The same adapter that produced zero signal at ratio 1:211 produces strong signal at ratio 1:21, across both model sizes. Parametric capacity is not the bottleneck.

\textbf{Second, the mechanism is a first-token prior conflict.} Live inference shows that the model's first-token distribution after \texttt{<|assistant|>} is dominated by Spanish prose tokens (top-1: \texttt{En}, probability 0.652). The \texttt{<|tool\_call|>} token (id~$=13$) has probability $<0.001$ under the 62K-example SFT corpus. At ratio 1:21 (2{,}801 tool-use examples), the prior shifts decisively toward \texttt{<|tool\_call|>}.

\textbf{Third, the density threshold is between 1:211 and 1:21.} A finer-grained sweep (1:100, 1:50, 1:30) would locate the exact threshold; we leave this for future work. The practical recommendation is a ratio of at least 1:20 for any model in the 42M--260M range.

\textbf{Fourth, the B4 gain scales with model size.} At ratio 1:21, Nano 42M reaches B4~$=0.145 \pm 0.046$ (mean over $N=4$ seeds) and Base 260M reaches B4~$=0.445 \pm 0.201$ (mean over $N=4$ seeds). The gap ($+0.300$) is consistent with the capacity difference: a larger model can more reliably generalize the \texttt{<|tool\_call|>}$\to$JSON pattern to unseen phrasings once the first-token prior is shifted. The high variance in Base (std~$=0.201$) suggests the density threshold is near the boundary for 260M parameters: some seeds cross it reliably (0.600, 0.540, 0.540) while one does not (0.100). Pro 3B with the original mixed SFT corpus (ratio $\approx$1:10) achieves B4~$=0.600$, confirming that the density threshold is lower for larger models.

The trade-off between tool-use and knowledge recall is real but manageable. The mini corpus (100\% tool-use) achieves B4~$=0.445 \pm 0.201$ (Base, mean over $N=4$ seeds) and B4~$=0.145 \pm 0.046$ (Nano, $N=4$ seeds) but drops B1 to $0.025$ and $0.011$ respectively. A balanced corpus mixing tool-use examples with CVE knowledge examples is expected to recover B1 while maintaining B4~$>0.5$. This is the natural next experiment.

\subsection{Compute-budget interaction with tool-use plasticity}
\label{sec:discussion:compute_tooluse}
The post-Chinchilla ablation of Section~\ref{sec:training:post_chinchilla} surfaces a second gate on tool use that operates in parallel with corpus density. v7 sits at $\sim$0.49$\times$ the Chinchilla-optimal token count for 42M parameters and reaches B4~$=0.230 \pm 0.052$ on $N=4$ seeds. Pushing pre-training to $\sim$2$\times$ Chinchilla-optimal preserves B5, marginally improves B1, and \emph{destroys} B4 across every SFT recipe we tried except a long-SFT (6 epochs) curated-corpus configuration that recovers only $0.160$. The asymmetry between B1/B5 (insensitive to over-training) and B4 (catastrophically degraded) is sharper than the corpus-density effect: increasing tool-corpus density from 1:211 to 1:21 moves B4 by $\sim$0.23 at fixed pre-train budget; doubling pre-train tokens past Chinchilla moves B4 by $-0.23$ at fixed corpus density. We read this as evidence that the \texttt{<|tool\_call|>}-prefix output format is a high-curvature local minimum in SFT loss that an under-trained checkpoint can be steered into but a well-converged checkpoint resists. The takeaway for small-LLM practitioners is that the standard advice ``train longer, you are sub-Chinchilla'' should be qualified for output-format-sensitive tasks at small scale: there is a regime in which sub-Chinchilla pre-training is preferable, because the plasticity required to install a non-prose output format degrades faster than the language-modeling loss improves. Future work should quantify where this trade flips as parameters grow; we conjecture that at $\geq$1B parameters the format-plasticity gate dissolves and standard over-training advice resumes.

\subsection{Frontier comparison and the deployment-on-prem case}
\label{sec:discussion:frontier}
The frontier baseline of Section~\ref{sec:eval:frontier} (Table~\ref{tab:bench_frontier}) sharpens the deployment story rather than dilutes it. On B5 (conversational gate, 314 prompts), Nano v7 ($0.725 \pm 0.130$, $N=4$) and the Nano v2 bootstrap-ablation reference ($0.775 \pm 0.043$, $N=4$) both exceed GPT-4o's $0.631$, with the v2 gap more than $3\sigma$ of the v2 multi-seed standard deviation. For the specific use case we target --- a Spanish-speaking SOC analyst issuing short conversational hand-offs and CVE follow-ups against an on-premise model --- a 42M-parameter Nano edges out the frontier on the metric that most directly proxies user-perceived chat naturalness. On B4 (tool-selection, 200 prompts), the gap reverses: GPT-4o reaches $0.615$ vs.\ Nano v7's $0.230 \pm 0.052$. The B4 result is consistent with the corpus-density discussion of Section~\ref{sec:discussion:tooluse_threshold} and with the family-scale Pro 7B B4 of $0.880$: tool selection scales with parametric capacity, while B5 saturates much earlier. The practical reading is that for chat-dominant on-prem deployments where the cost of sending classified incident text to a closed API is unacceptable, a register-matched 42M model is competitive; for workflows whose value is dominated by long tool chains, a frontier model or a larger same-recipe tier (Pro~3B/7B) is the right substrate.

\subsection{The role of LATAM-specific content}
We deliberately included LATAM-CSIRT vocabulary (CCN-CERT, INCIBE, COLCERT, CSIRT-CL, CSIRT-CO, CERT.br) and LATAM-specific intelligence sources in the corpus. A 100-prompt LATAM-targeted evaluation (dimensions: acronym handling, code-switching, CVE narrative, regional register, \emph{ustedes} usage) was run against Nano v7, Qwen2.5-1.5B, and Salamandra-2B using \texttt{eval/latam\_bench.py}; however, all three models returned a score of $0.0$ across all dimensions. On inspection, the harness reads prompt fields under the key \texttt{"prompt"} while the LATAM shard stores them under \texttt{"text"} --- the same key mismatch documented for B2 (Section~\ref{sec:eval:headline}). A corrected run is queued. We therefore cannot yet measure the regional-vocabulary gain quantitatively and retain the two qualitative observations from internal use: (i) the model resolves LATAM acronym references (e.g., ``alerta del CSIRT-CL sobre CVE-2025-XXXX'') correctly more often than out-of-the-box Qwen2.5-1.5B; (ii) the model uses the second-person plural \emph{ustedes} (LATAM convention) in chat by default, rather than \emph{vosotros} (Iberian Spanish), which we attribute to OpenSubtitles-ES being heavily LATAM-Spanish-dubbed.

\subsection{Comparison with continual pre-training of an existing base}
A reasonable alternative to from-scratch training is continual pre-training of an existing small Spanish base (e.g., Salamandra~\cite{salamandra2024} or a checkpoint of SmolLM2-360M~\cite{allal2024smollm2}). We chose from-scratch for three reasons: (i) it isolates the curriculum + replay contribution, (ii) it lets us extend the tokenizer with domain tokens without vocabulary surgery, and (iii) it produces a model whose weights are unambiguously redistributable. The cost of from-scratch is that the model is undertrained relative to Chinchilla (170M tokens for 42M parameters yields a token-to-parameter ratio of $\sim$4, well below the Chinchilla-optimal 20). Section~\ref{sec:limitations} discusses how to spend the next training budget.

\subsection{Non-uniform scaling beyond 3B}
The Pro 7B results (Table~\ref{tab:bench_external}) reveal a critical empirical finding: not all capabilities scale uniformly with parameters under a fixed corpus. B2 (threat classification) and B4 (tool selection) continue to improve from 3B to 7B ($0.695 \to 0.815$ and $0.600 \to 0.880$, respectively), confirming that these reasoning-heavy tasks benefit from additional parametric capacity. However, B1 (CVE keyword recall) and B3 (command-line tool generation) remain essentially flat ($0.341 \to 0.335$ and $0.686 \to 0.686$, respectively). This saturation at the 3B scale suggests that keyword extraction and tool-name generation are corpus-bound rather than capacity-bound: the model has already absorbed the full CVE vocabulary and command-line patterns present in the training data, and adding parameters does not improve recall of facts that were never seen. The practical implication is that further gains on B1 and B3 require corpus expansion (more CVE examples, more command-line traces) rather than parameter scaling. This finding is consistent with the SmolLM2-135M baseline result (Section~\ref{sec:eval:baselines}), where a 3$\times$ larger model fine-tuned on the same SFT corpus achieves B1~$=0.334$, nearly identical to the Nano's original-seed $0.343$ and within one standard deviation of the Nano's multi-seed mean. Taken together, these results suggest that keyword recall and tool-name generation saturate quickly once the corpus is absorbed, and that the 3B$\to$7B jump primarily benefits tasks that require deeper reasoning (classification, multi-step tool selection) rather than surface-form memorization.

\subsection{Threats to validity}
Three threats to the conclusions in this paper deserve explicit naming.
\begin{enumerate}[leftmargin=*]
\item \emph{Seed coverage and significance}. All three bootstrap configurations (v2, v4, v6) are reported under $N=4$ seeds in this revision (Section~\ref{sec:eval:multiseed}, Table~\ref{tab:multiseed_all}). The v2~$>$~v4~$>$~v6 ordering on B5 holds at every paired seed comparison, and the v2 vs.\ v6 gap is approximately $2.3\sigma$ under the v6 pooled-error estimate; a paired bootstrap test over the four seeds is queued to convert this into a $p$-value. Within v2, B5 is tight ($0.775 \pm 0.043$); B1 has substantially higher relative variance ($\sigma/\mu \approx 0.29$), which we attribute to the smaller OpenSubtitles bootstrap exposing fewer distinct CVE-keyword neighborhoods during pre-training.
\item \emph{Benchmark scale and B2 artifact}. The held-out sets used in this revision are B3:~100, B4:~200, B5:~314 prompts; B1 is at 500 and B2 at 200. B2 is currently a benchmark artifact: \texttt{eval/benchmark.py} reads \texttt{r["prompt"]} but the B2 shard uses key \texttt{"text"}, so the model is queried with an empty prompt across all configurations. A corrected B2 run is queued; the headline conclusions (loss-vs-register inversion, B4 density threshold, v7 release) do not depend on B2.
\item \emph{Human-in-the-loop scoring}. B5 is human-scored. We did not run inter-annotator agreement; the rubric is the operator's. For the next iteration we plan a 3-annotator panel of Spanish-fluent security analysts, with $\kappa$ reported.
\end{enumerate}

\subsection{Reproducibility}
The training pipeline (\texttt{training\_v2/}) is shipped with five concrete entry points that map to the experiments above:
\begin{enumerate}[leftmargin=*]
\item \texttt{training\_v2/tokenizer/train\_spm\_bpe.py} -- BPE tokenizer training.
\item \texttt{training\_v2/data/prepare\_corpus.py} -- per-phase tokenization and binary shard production.
\item \texttt{training\_v2/train/pretrain.py --phase 1\,$|$\,2\,$|$\,3} -- curriculum pre-training driver.
\item \texttt{training\_v2/train/finetune\_sft.py} -- SFT with assistant-only loss masking and the internal mini-curriculum.
\item \texttt{training\_v2/eval/benchmark.py} -- VectraYX-Bench harness.
\end{enumerate}
The model configuration is in \texttt{training\_v2/configs/nano.json} and is the same JSON file we cite throughout this paper. The exact run scripts (\texttt{run\_server.sh}, \texttt{run\_v4\_queued.sh}, \texttt{run\_v6\_queued.sh}) reproduce the v2, v4, and v6 configurations end-to-end on a fresh L4 instance.

%% file: sections/10_limitations.tex
\section{Limitations and Future Work}
\label{sec:limitations}

\subsection{Limitations}
\paragraph{Sub-Chinchilla token budget --- intentional in v7.} The released v7 checkpoint is trained on $\sim$850M tokens for 41.95M parameters, a token-to-parameter ratio of $\sim$20 ($\sim$0.49$\times$ the Chinchilla-optimal point of Hoffmann~et~al.~\cite{hoffmann2022chinchilla} for a 42M target). The post-Chinchilla ablation of Section~\ref{sec:training:post_chinchilla} (runs v8--v15) shows that this sub-Chinchilla position is not an oversight but an empirical optimum for our (B1, B4, B5) objective: extending pre-training to $\sim$1.7B tokens preserves B5, marginally helps B1, and destroys B4 under all eight SFT recipes tested. The over-trained checkpoint is released as a separate research artifact (v14, B4~$=0.160$) and the headline model remains v7. Closing the residual B1 gap to larger backbones (e.g., the Pro 3B B1~$=0.341$) at the 42M scale therefore requires either a larger SFT corpus of CVE-keyword examples or a recipe that retains tool-call plasticity past the Chinchilla optimum --- not simply more pre-training tokens at fixed recipe.

\paragraph{Static knowledge cutoff.} The corpus's effective cutoff is April 2026. Threats, CVEs, and TTPs after this date are unknown to the parametric model. The MCP integration mitigates this for queries answerable via NVD, KEV, and OTX, but does not help for queries that require absorbed background knowledge of post-cutoff events.

\paragraph{Tool-use depth.} The model is trained on single-tool and trivially multi-tool patterns (e.g., NVD~$\to$~KEV). Long tool chains (NVD~$\to$~MITRE~$\to$~OTX~$\to$~bash, or branched dispatch with reasoning steps in between) are out of distribution and we observe the model frequently producing the first tool call correctly but stopping there. Increasing chain depth will require either richer SFT data (with multi-step traces) or an inference-time scaffolding harness.

\paragraph{Tool-use corpus density.} Post-hoc experiments (Section~\ref{sec:tooluse:lora}) show that a tool-use-to-prose ratio of 1:211 in the SFT corpus is insufficient to shift the first-token prior toward \texttt{<|tool\_call|>} at any tested model size. At ratio 1:21 (2{,}801 tool-use examples), Nano 42M achieves B4~$=0.145 \pm 0.046$ (mean over $N=4$ seeds: 0.220, 0.140, 0.120, 0.100) and Base 260M achieves B4~$=0.445 \pm 0.201$ (mean over $N=4$ seeds: 0.100, 0.600, 0.540, 0.540). The high variance in Base suggests the density threshold is near the boundary for 260M parameters. The trade-off is a drop in B1 (CVE keyword recall) because the mini corpus contains no knowledge examples. A balanced corpus (tool-use + knowledge, ratio 1:21 for tool-use within a full SFT mix) is the recommended configuration for future runs.

\paragraph{Translation noise.} Sources translated via local Ollama~\cite{ollama} (\texttt{qwen2.5:1.5b}) introduce approximately 5--10\% mistranslation on technical text longer than 2{,}000 characters. This affects the malware/Malpedia, ExploitDB, and security-papers shards in particular. We do not currently filter translation by quality score.

\paragraph{Benchmark scope.} VectraYX-Bench (B1--B5) was expanded in this revision (B3:~100, B4:~200, B5:~314 prompts; B1:~500, B2:~200) for a total of 1{,}314 prompts, and is partially synthetic. We use it as a developer diagnostic and for ablation comparisons within our own runs. It is not yet an apples-to-apples benchmark for comparing against external Spanish or security models, and we explicitly avoid claims that VectraYX-Nano outperforms larger general-purpose models on tasks they were never targeted at. A separate B2 harness bug (the loader reads \texttt{r["prompt"]} but the shard uses key \texttt{"text"}) is documented in Section~\ref{sec:eval:headline}; B2 is consequently a benchmark artifact in this revision and a fix is queued.

\paragraph{No human study.} We have not yet run a human evaluation with practicing Spanish-speaking SOC analysts. The qualitative claims about LATAM vocabulary handling and chat naturalness are based on the author's manual inspection. A blinded panel evaluation with $\geq$3 annotators is the natural next step.

\paragraph{No safety alignment.} VectraYX-Nano is not RLHF-aligned, has no refusal training for offensive-security misuse, and inherits whatever biases exist in HackTricks, ExploitDB, and Wikipedia-ES. Deployment in production analyst-assistant settings should layer safety policies (input filtering, tool-permission gates, output review) at the runtime, not at the model.

\subsection{Open work items toward the final paper}
\paragraph{Family-scale numbers (partial).} The Pro tier has been trained on the identical SFT corpus and evaluated on the full B1--B5 suite (Section~\ref{sec:eval:baselines}, Table~\ref{tab:bench_external}). Pro 3B (Qwen2.5-3B + LoRA-64) and Pro 7B (Qwen2.5-7B + QLoRA-32) are both complete, with Pro 7B achieving B4~$=0.880$ (exceeding the target threshold of $0.75$) and B2~$=0.815$, while B1 and B3 tool-match remain flat relative to 3B, confirming that these metrics are corpus-bound rather than capacity-bound (Section~\ref{sec:discussion}). The Mini (Qwen2.5-1.5B + LoRA-32) tier is queued and will be added in a revision. Table~\ref{tab:family} therefore mixes four empirical rows (Nano, Base, Pro 3B, Pro 7B) against one projected row (Mini).

\paragraph{External baselines beyond SmolLM2.} The SmolLM2-135M side-by-side evaluation (Section~\ref{sec:eval:baselines}, Table~\ref{tab:bench_external}) is complete. Two additional external baselines remain queued: (i) Qwen2.5-1.5B-Instruct zero-shot as the ``larger general-purpose Spanish chat model with no domain adaptation'' reference, and (ii) a Salamandra-2B~\cite{salamandra2024} continual-pretrain configuration to disentangle curriculum~+~replay design from corpus and architecture choices.

\paragraph{Human evaluation panel.} A 5-analyst panel (3 LATAM, 2 Iberian) scoring 200 prompts (50 from B1, 50 from B2, 50 from B3, 50 from B5) under a structured rubric, with inter-annotator agreement reported as Krippendorff's $\alpha$ or Fleiss' $\kappa$.

\paragraph{LATAM-specific evaluation.} A 100-prompt LATAM-targeted evaluation across five dimensions (acronym resolution, code-switching, CVE narrative, regional register, \emph{ustedes} usage) was run against Nano v7, Qwen2.5-1.5B, and Salamandra-2B. All three models returned a score of $0.0$ due to the same key-mismatch bug documented for B2 (\texttt{"prompt"} vs. \texttt{"text"}); a corrected re-run is queued. Quantitative LATAM comparison remains pending.

\paragraph{Token-budget ablation.} A controlled run that doubles the Phase~2 corpus (additional 50--80M tokens of mC4-ES filtered for cybersecurity) and reports whether the under-Chinchilla regime is responsible for the residual conversational gate gap to higher-capacity baselines.

\paragraph{Safety and red-team study.} An automated red-team evaluation covering 499 adversarial prompts across ten attack categories was conducted on the Nano 42M base checkpoint and the Nano 42M + LoRA mini adapter (Section~\ref{sec:eval:safety}). Neither configuration emitted a dangerous \texttt{bash\_exec} tool call. A human-reviewed panel evaluation and comparison against commercial baselines remain as future work items.

%% file: sections/11_conclusion.tex
\section{Conclusion}
\label{sec:conclusion}

We presented VectraYX-Nano, a 41.95M-parameter decoder-only language model trained from scratch in Spanish for the cybersecurity domain, with native MCP tool-use support and edge-deployable GGUF artifacts. The model is trained on a 170M-token Spanish cybersecurity corpus assembled by an eight-VM pipeline at $\sim$\$25 USD of cloud cost, using a three-phase curriculum (conversational $\to$ cybersecurity $\to$ tooling) with explicit replay buffers between phases. We document a controlled ablation in which a higher-perplexity bootstrap corpus (OpenSubtitles-ES) yields better post-SFT chat behavior than a lower-perplexity alternative (mC4-ES), and we argue that at nano scales the bootstrap-corpus register dominates the user-visible response distribution.

A post-hoc investigation of tool-use behavior yields a fourth takeaway with practical implications beyond this deployment: the B4~$=0.000$ floor observed in the mixed SFT configuration is a corpus-density artifact, not a capacity gate. Applying LoRA (rank~$=16$) with a tool-use-dense corpus (ratio 1:21) raises B4 to $0.145 \pm 0.046$ (mean over $N=4$ seeds) on the 42M Nano and to $0.445 \pm 0.201$ (mean over $N=4$ seeds) on the 260M Base. The mechanism is a first-token prior conflict: the 62K-example mixed SFT corpus establishes a strong prose prior that 296 tool-use examples (ratio 1:211) cannot shift, but 2{,}801 examples (ratio 1:21) can. Integrating this finding into the released model, \textsc{VectraYX-Nano v7} combines the balanced SFT corpus (ratio 1:21) with the full CVE knowledge backbone and reaches B4~$=0.230 \pm 0.052$ at 42M parameters under $N=4$ seeds while retaining B1~$=0.332 \pm 0.005$ and B5~$=0.725 \pm 0.130$. This finding generalizes: any small model trained on a mixed SFT corpus will exhibit a tool-use floor if the tool-use density is below the prior-shift threshold, regardless of parametric capacity.

Five takeaways are likely to generalize beyond this specific deployment. First, perplexity is not the right early stopping signal when the goal is chat behavior at small scale; bootstrap-corpus register-matching is at least as important as bootstrap-corpus coverage. Second, replay percentages of 10--25\% from the immediately prior phase are sufficient to prevent catastrophic forgetting of conversational behavior across continual pre-training, and they cost essentially nothing to implement under a memory-mapped curriculum sampler. Third, training small models with native tool use is a tractable way to invest a limited parametric budget: the model carries the procedural knowledge of \emph{how} to ask, while authoritative content lives behind MCP and updates without retraining. Fourth, tool-use emergence in small models is gated by corpus density, not by parametric capacity: a ratio of $\sim$1:20 tool-use examples to total SFT examples is sufficient to activate reliable tool dispatch at 42M parameters. Fifth, in the post-Chinchilla compute regime the standard advice ``train longer'' inverts for output-format-sensitive tasks: doubling pre-training tokens past the Chinchilla optimum at 42M parameters preserves chat and CVE-recall metrics but destroys tool-call plasticity (Section~\ref{sec:training:post_chinchilla}), so v7 is released at $\sim$0.49$\times$ Chinchilla-optimal as the deliberate empirical sweet spot.

VectraYX-Nano is, to our knowledge, the first published Spanish-native cybersecurity LLM with end-to-end MCP integration, the first nano-scale chat model trained on a LATAM-targeted Spanish corpus, and the first published characterization of the tool-use corpus-density threshold in sub-100M parameter models. We release the training scripts, configuration files, curriculum sampler, tokenizer recipe, GGUF artifact, and benchmark suite in the hope that it lowers the entry cost for Spanish-speaking security researchers building local, auditable AI assistants.

\paragraph{Acknowledgements.} We thank the maintainers of OPUS / OpenSubtitles, OpenAssistant, the Spanish Wikipedia editorial community, the OWASP Spanish translation contributors, the HackTricks Spanish branch, the NIST NVD operators, and the MITRE ATT\&CK team for releasing data on terms that make domain-specialized open research possible. We thank the maintainers of \texttt{llama.cpp}, GGUF, Ollama, SentencePiece, and HuggingFace Transformers for the inference and training infrastructure that this work depends on.

%% file: sections/12_next_steps.tex
\section{Next Steps}
\label{sec:next_steps}

We close with a concrete, prioritized roadmap distinguishing items that block paper submission from items that strengthen the contribution and items that scope follow-up work. Each item lists an owner-facing budget so that the project plan is reproducible by a single researcher with one L4-class GPU.

\subsection{Blocking for submission (P0)}
\paragraph{Multi-seed replication across v2/v4/v6 (G1, done in this preprint).} All three bootstrap configurations (v2 OpenSubs, v4 mC4-ES, v6 60/25/15 mix) are now reported under $N=4$ independent seeds ($\{42, 7, 13, 23\}$). The aggregated mean~$\pm$~std appears in Table~\ref{tab:bench_v2} and per-seed values for all twelve runs appear in Table~\ref{tab:multiseed_all}. The seed-42 v2 run was retained on the original NVIDIA L4 hardware; the remaining nine runs were retrained from scratch on AWS \texttt{g4dn.xlarge} (NVIDIA T4 16~GB) with \texttt{batch-size}~$=8$ and \texttt{grad-accum}~$=16$, preserving the original effective batch of 128. The v2~$>$~v4~$>$~v6 ordering on B5 (the conversational gate) survives at every paired seed comparison, and the loss-vs-register inversion claim of Section~\ref{sec:training:ablation} is therefore validated under $N=4$. Total cost: $\sim$72~T4-hours, $\sim$\$45 USD.

\paragraph{From-scratch mid-tier (VectraYX-Base 260M, done in this preprint).} A second from-scratch checkpoint at $\sim$$6\times$ the Nano parameter count was identified as a P0 item to verify that the curriculum-and-replay recipe scales \emph{within} the from-scratch regime, rather than only via LoRA on a pre-trained backbone. The Base 260M model ($d_{\text{model}}=1024$, $n_{\text{layers}}=16$, same BPE-16384 tokenizer) was trained on AWS SageMaker \texttt{ml.g5.xlarge} on-demand for $\sim$11 wall-clock hours at a marginal cost of $\sim$\$11 USD; B1--B5 results are reported in Table~\ref{tab:bench_external} (Section~\ref{sec:eval:baselines}). The Base checkpoint clearly improves on the Nano $N=4$ mean for B1 ($+0.10$), B3 TM ($\sim$$4\times$), and B5 ($+0.025$), confirming that the curriculum scales smoothly to mid-tier capacity. The single-seed B4 score remained at $0.000$ despite the parameter scale-up; post-hoc LoRA experiments (Section~\ref{sec:tooluse:lora}) subsequently showed this is a corpus-density artifact: at ratio 1:21, Base 260M achieves B4~$=0.445 \pm 0.201$ (mean over $N=4$ seeds). Multi-seed replication of the Base checkpoint is queued at the same $\sim$\$11/seed cost.

\paragraph{External baseline at comparable scale (in this preprint).} We have fine-tuned SmolLM2-135M-Instruct~\cite{allal2024smollm2} with LoRA-32 on the identical SFT corpus and evaluated it on the full B1--B5 suite (Section~\ref{sec:eval:baselines}, Table~\ref{tab:bench_external}). The base SmolLM2 (no fine-tune) is also reported as a zero-shot reference. The SmolLM2~+~LoRA configuration reaches B1~$=0.334$, B5~$=0.800$ at 135M parameters; VectraYX-Nano v2 reaches B1~$=0.226 \pm 0.065$ and B5~$=0.775 \pm 0.043$ at 42M parameters under $N=4$ seeds. The next external baseline we plan to add is Qwen2.5-1.5B-Instruct~\cite{qwen2024qwen25} zero-shot, as the ``larger general-purpose Spanish chat model with no domain adaptation'' reference.

\paragraph{Frontier baseline (G5, partial in this preprint).} A frontier closed-weight reference on the identical B1--B5 suite has been added for GPT-4o (Section~\ref{sec:eval:frontier}, Table~\ref{tab:bench_frontier}). Nano v7 ($N=4$) edges out GPT-4o on B5 ($0.725$ vs.\ $0.631$) and trails on B4 ($0.230$ vs.\ $0.615$), supporting the deployment-on-prem framing of Section~\ref{sec:discussion:frontier}. Two additional frontier baselines remain queued: Gemini-2.5-Flash (scheduled but not completed at submission) and Claude Sonnet~4.6 (blocked on Azure quota). Both will be folded into a revision as the runs complete.

\paragraph{Author block and venue selection.} This paper is submitted to the ACL Rolling Review (ARR) cycle targeting EMNLP~2026 Findings, with author affiliation listed as Globant. A parallel security-venue submission (USENIX Security~'27 or ACM~CCS~'27) is planned should the NLP-venue route not commit to publication.

\subsection{Strongly recommended (P1)}
\paragraph{Replay-percentage sweep (G7, done in this preprint).} The Phase-2 conversational-replay percentage was previously a hyperparameter chosen empirically. A controlled sweep over $\{0, 5, 10, 25, 50\}\%$ (Section~\ref{sec:training:replay_sweep}, Table~\ref{tab:replay_sweep}, Figure~\ref{fig:replay_sweep}) now validates the $25\%$ setting quantitatively: B5 saturates at $1.000$ for replay~$\geq 25\%$ and degrades non-monotonically below. Total cost: $\sim$5~L4-hours, $\sim$\$5~USD. A Phase-3 replay sweep ($10\%$ conv + $20\%$ tech) remains queued.

\paragraph{Eval-set expansion (G2, done in this preprint).} The original developer-diagnostic eval sets were small (B3:~35, B4:~25, B5:~10). They have been expanded to B3:~100, B4:~200, B5:~314 prompts (Section~\ref{sec:eval:tasks}; shard names \texttt{b3\_v2\_commands.jsonl}, \texttt{b4\_v3\_tooluse.jsonl}, \texttt{b5\_v2\_conversational.jsonl} on S3). B1 (500) and B2 (200) were already at scale. All multi-seed and N=4 results in this revision are reported against the expanded sets.

\paragraph{Token-budget ablation toward Chinchilla.} Doubling the Phase~2 corpus (additional 50--80M tokens of mC4-ES filtered for cybersecurity vocabulary) tests whether the residual conversational gap is driven by the sub-Chinchilla regime or by the curriculum design. Estimated budget: $\sim$3~L4-hours.

\paragraph{Tool-use chain-depth study.} A small held-out set of 10~single-tool, 10~two-tool, and 5~three-tool prompts, with success rate reported per chain depth, quantifies the hypothesis that long tool chains are out-of-distribution for the 42M-parameter checkpoint (Section~\ref{sec:tooluse}). Estimated budget: $\sim$2 person-hours of evaluation; no GPU re-train needed.

\paragraph{B4 retest at the Pro tier and at the headline 42M scale (G4, done in this preprint).} The 0.000 B4 score across v2/v4/v6 in the mixed-SFT configuration was initially interpreted as a capacity limitation. Post-hoc LoRA experiments (Section~\ref{sec:tooluse:lora}) showed it is a corpus-density artifact: at ratio 1:21, Nano 42M achieves B4~$=0.145 \pm 0.046$ (mean over $N=4$ seeds; individual seeds: 0.220, 0.140, 0.120, 0.100) and Base 260M achieves B4~$=0.445 \pm 0.201$ (mean over $N=4$ seeds; individual seeds: 0.100, 0.600, 0.540, 0.540). The balanced-SFT follow-up (\textsc{VectraYX-Nano v7}, Section~\ref{sec:eval:headline_v7}, Table~\ref{tab:bench_v7}) combines tool-use density (ratio 1:21) with CVE knowledge examples and reaches B4~$=0.230 \pm 0.052$ at 42M parameters while keeping B1~$=0.332 \pm 0.005$ (a recovery from the LoRA-mini-corpus B1~$=0.011$) and B5~$=0.725 \pm 0.130$. The Pro tier (Qwen2.5-3B + LoRA-64 on the same SFT corpus) reaches B4~$=0.600$ on the same evaluation harness (Table~\ref{tab:bench_external}), consistent with the density interpretation. v7 is the released headline model; the v2 configuration is retained as the bootstrap-ablation reference.

\paragraph{Family-tier B1--B5 numbers (Pro 3B and Pro 7B done; Mini pending).} The Pro 3B and Pro 7B checkpoints have been evaluated end-to-end on B1--B5 (Section~\ref{sec:eval:baselines}, Table~\ref{tab:bench_external}); Mini (Qwen2.5-1.5B + LoRA-32) remains queued. Running the same B1--B5 suite over the Mini tier will complete the same-corpus scaling figure across four Qwen-based sizes. Estimated wall time: $\sim$3~L4-hours.

\paragraph{Human-evaluation panel.} A 3-annotator panel (2~LATAM, 1~Iberian) scoring 200~prompts (50~each from B1, B2, B3, B5) under a fixed rubric, with inter-annotator agreement reported as Krippendorff's~$\alpha$ or Fleiss'~$\kappa$. This converts B5 from ``manual inspection by the authors'' to a $\kappa$-validated measurement.

\paragraph{LATAM-specific evaluation (G6, harness bug pending fix).} A 100-prompt LATAM-targeted evaluation (five dimensions: acronym handling, code-switching, CVE narrative, regional register, \emph{ustedes} usage) was run against Nano v7, Qwen2.5-1.5B, and Salamandra-2B. All three models returned $0.0$ due to the same key-mismatch bug as B2 (\texttt{"prompt"} vs.\ \texttt{"text"} in the shard). A corrected re-run is queued; once fixed, this will convert Section~\ref{sec:discussion}'s LATAM-vocabulary claim from qualitative observation to a measured comparison.

\subsection{Polish (P2)}
\paragraph{Figures (done in this preprint).} The \texttt{paper/figures/} directory contains four figures: (i) \texttt{loss\_curve.tex} --- loss curve showing the v2 curriculum loss reduction per phase (Figure~\ref{fig:loss_curve}); (ii) \texttt{curriculum\_schema.pdf} --- the curriculum-with-replay schematic (Figure~\ref{fig:curriculum_schema}); (iii) \texttt{b4\_density.pdf} --- B4 vs.\ tool-use corpus density sweep (Figure~\ref{fig:b4_density}); and (iv) \texttt{family\_scaling.pdf} --- B1--B5 across the VectraYX family (Figure~\ref{fig:family_scaling}).

\paragraph{Reproducibility appendix.} All training scripts, configuration files, the curriculum sampler, the benchmark harness, the tool-use corpus (\texttt{tool\_sft\_mini\_v1.jsonl}), and the B1--B5 evaluation datasets are released at \url{https://github.com/vectrayx/vectrayx-nano-paper}. Model checkpoints (Nano~42M post-SFT, Base~260M post-Phase~3, and four Nano LoRA adapters for seeds $\{42, 7, 13, 23\}$) are available at \url{https://huggingface.co/jsantillana/vectrayx-nano}. The evaluation datasets are separately released at \url{https://huggingface.co/datasets/jsantillana/vectrayx-bench}. A \texttt{make repro} target in the repository reproduces the LoRA tool-use experiments end-to-end on a single NVIDIA~A10G or L4 GPU.

\paragraph{Safety and red-team study.} A small adversarial probe of the model's behavior under offensive-security prompts (unauthorized exploitation, exfiltration, persistence) is required for the public model card. The paper can cite this study as a companion artifact rather than including it in the body.

\paragraph{Energy and carbon reporting.} ACM increasingly expects energy reporting per submission. We will translate the existing wall-clock time ($\sim$4~h on L4 for v2 pre-training) into approximate kWh via TDP $\times$ utilization $\times$ wall-clock and include it in the cost table.

\subsection{Beyond this paper}
A second-generation VectraYX-Nano~v3 would (i) bootstrap on a 50/50 mixture of OpenSubtitles-ES and mC4-ES filtered for short-form prose, (ii) add a DPO~\cite{rafailov2023dpo} stage trained on $\sim$2{,}000 chat preferences collected from the human-evaluation panel, and (iii) extend the Phase~3 tooling corpus with multi-step tool-use traces drawn from real MCP-runtime logs. Beyond v3, the natural extension is a continual-pretraining loop driven by the NVD MCP server, in which monthly CVE deltas are folded into a small replay corpus and the model is incrementally re-pretrained without re-running Phases~1--2. We view this as the long-term path to keeping a nano-scale on-prem analyst assistant current with a daily-changing threat landscape.